\documentclass[10pt,twocolumn,letterpaper]{article} 

\usepackage[pagenumbers]{cvpr}

\providecommand{\floored}{\rlap{$^{\dagger}$}}

\def\rvepsilon{{\boldsymbol{\epsilon}}}

\def\rvc{{\mathbf{c}}}

\def\rvu{{\mathbf{u}}}
\def\rvv{{\mathbf{v}}}

\def\rvx{{\mathbf{x}}}

\def\rvz{{\mathbf{z}}}

\usepackage{colortbl}
\definecolor{tintbase}{HTML}{EB6834}

\usepackage{algorithm}
\usepackage{algpseudocode}

\definecolor{cvprblue}{rgb}{0.21,0.49,0.74}
\usepackage[pagebackref,breaklinks,colorlinks,allcolors=cvprblue]{hyperref}

\def\confName{CVPR}
\def\confYear{2027}

\title{Less Noise, More Motion: Revisiting Human Motion Diffusion}

\author{Setareh Cohan \qquad Michiel van de Panne\\
University of British Columbia\\
Vancouver, Canada\\
{\tt\small \{setarehc, van\}@cs.ubc.ca}
}

\begin{document}
\newsavebox{\teasermotions}
\newcommand{\teaserpanel}[1]{\parbox[b][\ht\teasermotions][c]{\linewidth}{\centering\includegraphics[width=\linewidth]{#1}}}
\twocolumn[{%
  \renewcommand\twocolumn[1][]{#1}%
  \maketitle
  \begin{center}
    \captionsetup{type=figure}
    \includegraphics[width=\textwidth]{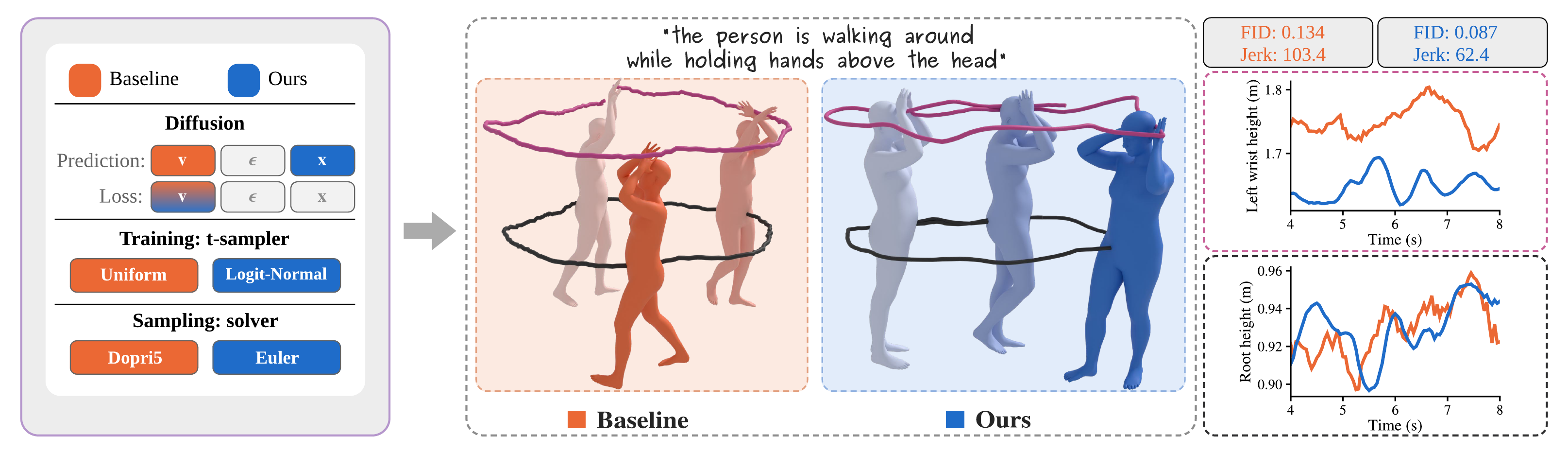}
    \caption{\textbf{A simple recipe for diffusion on raw joint positions.}
    Left:~The design choices we study. Orange marks the standard flow matching baseline ($\rvv$-prediction, $\rvv$-loss) and blue marks ours ($\rvx$-prediction,
    $\rvv$-loss). The baseline (\cref{sec:setup}) samples $t$ uniformly and uses its stronger but costlier adaptive solver (Dopri5). Ours uses a logit-normal sampler for $t$ and a simple Euler solver.
    Middle:~Motions generated by each recipe for the same prompt.
    Right:~Height of the root joint (pelvis) and left wrist over four seconds of the two generated motions. The jitter in baseline shows as frame-to-frame fluctuations in the curves. Boxes give the FID and Jerk of each model.}
    \label{fig:teaser}
  \end{center}%
}]

\begin{abstract}
    Diffusion and flow matching human motion models require three coupled design choices: the motion representation (raw joint positions, engineered redundant features, or autoencoder latents), the prediction target (velocity, noise, or clean motion), and the loss space. 
    Current works inherit their prediction targets and loss objectives from image generation or set them for convenience and rely on redundant or latent representations, which can help hide unwanted generative noise at the cost of error-prone processing, lossy reconstruction, and challenges in applying spatial control. Raw joint positions are a simpler representation, but we observe that standard velocity and noise prediction models operating on them produce motion jitter and bone-length fluctuations that metrics such as FID fail to capture. We conduct a systematic study on this design space and train a flow-matching Transformer with all nine prediction-loss combinations on the HumanML3D and KIT-ML datasets. We find that clean motion prediction is key to producing high-quality predictions for diffusion models trained on raw joint position data. This matches the FID of velocity prediction with half the Jerk (a proxy for jitter), converges in half the training iterations, and performs well even at 5M parameters. When combined with a logit-normal timestep sampler and simple Euler solver, it lowers HumanML3D FID by 40\% over prior raw-motion diffusion methods at three to four times fewer network evaluations. This advantage vanishes when applied to autoencoder latent motion features, which supports our hypothesis that the relative advantage of clean motion prediction stems from the intrinsic low-dimensional nature of plausible human motion data. \url{https://setarehc.github.io/LessNoiseMoreMotion/}
\end{abstract}

\section{Introduction}
\label{sec:introduction}
Diffusion~\cite{sohl2015deep, ho2020denoising, song2020score} and flow matching models~\cite{lipman2022flow, liu2022flow} have emerged as a powerful class of generative models and achieved state-of-the-art (SOTA) performance in different domains such as image, audio, and video generation. Their high sample quality, multimodal distribution coverage, and flexible editing capabilities make them an ideal framework for human motion generation~\cite{tevet2022human, zhang2024motiondiffuse, kim2023flame} and editing~\cite{karunratanakul2023guided, xie2024omnicontrol, cohan2024flexible}, and they are successfully deployed in a wide range of applications such as human motion modeling for games, virtual reality, simulations, and robotics. As they gain more use, improving them in terms of distributional fidelity, motion quality and plausibility, generation speed, and conditioning is of great interest.

Despite their remarkable success, diffusion motion models continue to struggle with high-frequency jitter and temporal inconsistencies \cite{chen2025free, cuba2025flowmotion}. This inherent generative noise has been largely masked by architectural choices and highly engineered motion representations. For instance, the widely used HumanML3D~\cite{guo2022generating} representation injects redundancies such as joint velocities on top of relative-local joint positions and foot contacts that implicitly regularize the output, while latent diffusion models rely on pre-trained decoders that act as smoothing filters on the noisy latent predictions.

These structural guardrails have inherent limitations. Latent spaces introduce lossy reconstruction after compression, and require training a separate autoencoder~\cite{li2026back}. Redundant representations require error-prone pre and post processing~\cite{meng2025rethinking}, while relative-local representations complicate downstream conditioning on global targets~\cite{karunratanakul2023guided, xie2024omnicontrol} and keyframes~\cite{cohan2024flexible}. Following the broader trend of operating directly on pixels or raw data spaces~\cite{saharia2022photorealistic, hoogeboom2023simple}, training on raw joint positions is favorable for simplicity, scalability, and direct support of spatial motion control and editing tasks~\cite{meng2025rethinking, meng2025absolute}. Recent methods such as  ACMDM~\cite{meng2025absolute} embrace this paradigm and show that training directly on raw joint positions yields excellent results as measured by standard text-to-motion (T2M) metrics (\eg FID, R-Precision). 

However, relying exclusively on these quantitative metrics borrowed from the text-to-image domain, paints an incomplete picture of visual quality of human motion sequences. Through empirical evaluation, we observe that when training on raw joint positions, the underlying generative noise is re-exposed and a significant portion of samples demonstrate noticeable high-frequency jitter and unrealistic fluctuations, significantly compromising the visual and physical plausibility of the motions. Standard flow-matching or noise-prediction formulations, as used in ACMDM, predict the velocity or noise across a high-dimensional ambient space. Because human motion data is highly constrained by the kinematic tree, fixed bone lengths, and rotation limits, valid poses occupy a small, low-dimensional manifold within this space~\cite{safonova2004synthesizing, grochow2004style, wang2008gaussian, holden2015learning}. 
Without an engineered representation or geometric losses to constrain the solution space, small errors in velocity or noise prediction can push the trajectory off this manifold, resulting in spatial inconsistencies and visible jitter.

Recent studies in the image domain suggest that when operating on raw pixels, which also exhibit soft statistical regularities and are hypothesized to live on a low-dimensional manifold~\cite{chapelle2009semi,alemi2016deep}, directly predicting clean data naturally grounds the generation process and reduces off-manifold drift \cite{li2026back, jin2026revisiting}. We hypothesize that applying this clean data prediction to human motion generation will inherently anchor the spatial geometry and restore physical plausibility without requiring redundant representations, latent autoencoders (AEs) or auxiliary losses. Therefore, we conduct a systematic study of prediction targets (velocity, noise, data) and an analogous set of loss objectives across flow matching models. Using a self-contained Transformer architecture trained on raw joint positions, we evaluate all nine prediction-loss combinations on the standard HumanML3D~\cite{guo2022generating} and KIT-ML~\cite{plappert2016kit} motion capture (MoCap) datasets. 

Our work yields the following key findings:
\begin{itemize}
    \item {\textbf{The choice of prediction target matters far more than the choice of loss space.} In particular, predicting clean motions ($\mathbf{x}$-prediction) yields the highest distributional fidelity and the smoothest motion (least jitter).}
    \item {\textbf{Superior training and parameter efficiency:} $\mathbf{x}$-prediction achieves target quality in roughly half the training iterations of velocity ($\mathbf{v}$-prediction). A compact 5M-parameter $\mathbf{x}$-prediction model yields smoother motion than a 130M-parameter $\mathbf{v}$-prediction model.}
    \item {\textbf{Lower inference cost.} Combining $\mathbf{x}$-prediction with a logit-normal timestep sampler and a simple Euler sampler lowers FID on HumanML3D and halves jerk relative to the $\mathbf{v}$-prediction recipe of ACMDM, at a quarter of the sampling cost of its adaptive solver.}
    \item {\textbf{Latent space validates the manifold hypothesis.} With identical models using AE latents, the gap between $\mathbf{x}$-prediction and $\mathbf{v}$-prediction vanishes, albeit at the cost of other limitations. This confirms that the distinct advantage of $\mathbf{x}$-prediction is fundamentally tied to the low dimensionality of raw spatial data.}
\end{itemize}

\section{Related Work}
\label{sec:related_work}

\paragraph{Diffusion Models for Human Motion.}
Diffusion models were adapted to T2M generation by MDM~\cite{tevet2022human}, MotionDiffuse~\cite{zhang2024motiondiffuse} and FLAME~\cite{kim2023flame}, and have since become a key generative backbone for human motion modeling. Subsequent work has mostly focused on architecture improvements and a wider range of tasks, \eg latent diffusion for faster sampling~\cite{chen2023executing, dai2024motionlcm}, flow-based models for easier formulation and efficiency~\cite{hu2023motion, cuba2025flowmotion, meng2025absolute}, autoregressive variants for real-time synthesis~\cite{tevet2025closd, zhao2025dartcontrol, gou2025control}, spatial control and editing through guidance or conditioning~\cite{karunratanakul2023guided, xie2024omnicontrol, shafir2024human, cohan2024flexible}, and physics-based correction~\cite{yuan2023physdiff, watanabe2025simdiff}. Across this line of work, the diffusion formulation is inherited from the image domain or set for convenience: 
MDM predicts the clean motion, in part to enable the use of auxiliary geometric losses;
MotionDiffuse and MLD~\cite{chen2023executing} predict the noise; and ACMDM~\cite{meng2025absolute} adopts flow matching with velocity prediction. We hold the architecture fixed and systematically investigate the prediction-target and loss-space choices, and related choices such as the diffusion timestep sampler.

\paragraph{Human Motion Representations.}
Early human motion models encode motion as joint angle rotations along a kinematic tree~\cite{fragkiadaki2015recurrent, martinez2017human, loper2023smpl} strictly preserving bone lengths but suffering from error accumulation down the kinematic chain. Subsequent work adopted root-relative canonical coordinates~\cite{holden2016deep, ahuja2019language2pose, guo2020action2motion} and represented the root relative to the previous frame and each pose in local root coordinates. The dominant T2M representation~\cite{guo2022generating} adopts the same relative-local representation and augments it with redundant features such as joint velocities and foot contacts. The redundancy acts as an implicit regularizer and the root-relative encoding provides translation invariance. However, this requires error-prone conversions between representations~\cite{meng2025rethinking}, and conditioning on global control signals and keyframes requires representation conversions~\cite{karunratanakul2023guided, cohan2024flexible, tevet2025closd} or incorporation via additional learned components~\cite{xie2024omnicontrol}. Latent diffusion models~\cite{chen2023executing, uchida2025mola} instead compress motion with a learned autoencoder. The latent is compact and the decoder smooths the output, at the cost of designing and training a separate autoencoder whose reconstruction bounds the achievable quality~\cite{li2026back}. ACMDM~\cite{meng2025absolute} trains the same diffusion model on raw joint positions and their AE latents. On raw joint positions, it outperforms the absolute version of the redundant representation of \citet{guo2022generating} and has close performance to the latent version, which it keeps as its final model. We adopt the raw representation and use redundant and latent representations only as control experiments, which lets us isolate what these engineered redundant representations were masking.

\paragraph{Prediction Targets, Losses, and Diffusion on Raw Data.}
A diffusion or flow model can be trained to predict the noise~\cite{ho2020denoising}, the score~\cite{song2020score}, the clean sample~\cite{ramesh2022hierarchical}, or the velocity~\cite{salimans2022progressive, lipman2022flow, liu2022flow}. The targets are affine functions of one another at any timestep, so the choice only changes the errors the loss optimizes at each noise level without changing the model class~\cite{karras2022elucidating, gagneux2026training}. Common practice has settled on noise prediction for DDPMs~\cite{ho2020denoising} and velocity prediction for flow matching, and motion models have largely adopted these defaults. 
The question has resurfaced in the image domain with the return to raw pixel spaces. Pixel-space diffusion without a latent autoencoder~\cite{saharia2022photorealistic, hoogeboom2023simple} works best when the network predicts the clean image, which lies on a low-dimensional manifold, rather than the noise or velocity which do not~\cite{li2026back, jin2026revisiting}. Motion is a more extreme case than pixels: valid poses are confined to a low-dimensional manifold by hard kinematic constraints rather than by soft statistical regularities. This points to the need for an analogous understanding regarding the choice of prediction-targets and loss objectives for diffusion motion models. The notion of quality is also broader than in the image domain as generated motions should be smooth and physically plausible. We close this gap by evaluating all nine prediction-loss combinations on raw joint positions, and computing a number of physical plausibility metrics alongside the T2M metrics. Clean-motion prediction wins on every axis: it matches or improves FID while halving jerk (a proxy for visible jitter), converges in half the training iterations, and tolerates a cheaper fixed-step solver.

\paragraph{Jitter and Physical Plausibility.}
A generated motion should be judged not only by how well it matches the data distribution but also by whether it is physically plausible. Joints should move smoothly, bone lengths should remain fixed, and feet should not penetrate the floor. Diffusion models are known to violate these constraints with high-frequency jitter and temporal inconsistencies~\cite{curreli2025nonisotropic, chen2025free, cuba2025flowmotion}. Prior work has tried to mitigate this via additional geometrical loss terms on position, velocity and foot contacts~\cite{tevet2022human}, adding a frequency-domain consistency loss~\cite{chen2025free}, cleaning samples with physics simulators~\cite{yuan2023physdiff, watanabe2025simdiff}, and with latent models by relying on the decoder to smooth the output~\cite{chen2023executing}. 
We show that for raw joint position data, predicting clean motion removes most of the jitter and bone-length drift without requiring any auxiliary loss or the use of latent encodings. To better quantify jitter and physical plausibility,  
we report Jerk, Acceleration, Bone-length Change, and Floor-penetration metrics alongside T2M metrics (\eg FID) throughout. 

\section{Background}
\label{sec:background}

Diffusion models~\cite{ho2020denoising, song2020score} and flow matching~\cite{lipman2022flow, liu2022flow} are two formulations of the same family of generative models. They both train a network to gradually transform a simple distribution, typically Gaussian noise, into a target data distribution through iterative refinements, and are primarily different in the noise schedule and the weighting of the loss over diffusion steps $t$~\cite{gao2025diffusionmeetsflow, albergo2025stochastic}. This allows us to study a single paradigm without loss of generality, and we adopt the flow matching formulation~\cite{lipman2022flow,liu2022flow} with a linear schedule~\cite{lipman2022flow,liu2022flow,albergo2023building}.

\subsection{Flow Matching}
\label{sec:flow-matching}
Flow matching with a linear schedule operates on noisy samples:
\begin{align}
    \label{eq:flow-zt}
    \rvz_t  = t\,\rvx + (1-t)\,\rvepsilon,  
\end{align}
which moves from noise at $t = 0$ to data at $t = 1$ along a straight conditional path with flow velocity:
\begin{align}
    \label{eq:flow-v}
    \rvv = \rvx - \rvepsilon.
\end{align}

The network is trained with the flow matching loss of:
\begin{align}
    \label{eq:flow-loss}
    \mathcal{L} = \mathbb{E}_{t, \rvx, \rvepsilon} \big\|\rvv_\theta(\rvz_t, t) - \rvv \big\|^2.
\end{align}

Given the trained network $\rvv_\theta$, samples are generated by integrating the ordinary differential equation (ODE)~\cite{chen2018neural,song2020score,lipman2022flow, song2020denoising} $\mathrm{d}\rvz_t = \rvv_\theta\,\mathrm{d}t$ from $t=0$ to $t=1$, using a numerical solver (\eg Euler, Heun).

\begin{table}[t]
  \centering
  \small
  \setlength{\tabcolsep}{2pt}
  \renewcommand{\arraystretch}{1.45}
  \caption{Recovering each target (columns) from the network output prediction (rows) under the
  linear schedule of \cref{eq:flow-zt}. Tinted cells are direct network outputs.}
  \label{tab:conversions}
  \newcolumntype{C}{>{\centering\arraybackslash}p{0.30\linewidth}}
  \begin{tabular}{@{}r@{\hspace{4pt}}CCC@{}}
    \toprule
     & $\hat{\rvv}$ & $\hat{\rvepsilon}$ & $\hat{\rvx}$ \\
    \midrule
    $\rvv_\theta$ & \cellcolor{gray!14}$\rvv_\theta$ & $\rvz_t - t\,\rvv_\theta$ & $\rvz_t + (1-t)\,\rvv_\theta$ \\
    $\rvepsilon_\theta$ & $(\rvz_t - \rvepsilon_\theta)/t$ & \cellcolor{gray!14}$\rvepsilon_\theta$ & $(\rvz_t - (1-t)\,\rvepsilon_\theta)/t$ \\
    $\rvx_\theta$ & $(\rvx_\theta - \rvz_t)/(1-t)$ & $(\rvz_t - t\,\rvx_\theta)/(1-t)$ & \cellcolor{gray!14}$\rvx_\theta$ \\
    \bottomrule
  \end{tabular}
\end{table}

\subsection{Prediction Output and Loss Space}
\label{sec:prediction-loss}

\paragraph{Prediction Output.}
\label{par:prediction}
In flow matching, the network can output any of the three targets $\rvv$, $\rvepsilon$, or $\rvx$. 
At any diffusion step $t$, the network prediction output (either $\rvv_\theta$, $\rvepsilon_\theta$, or $\rvx_\theta$) and the noisy sample ($\rvz_t$) are known, while the other two targets are unknown. Using the linear constraints of \cref{eq:flow-zt} and \cref{eq:flow-v}, the system of equations can be solved to derive the unknowns. For example, with $\rvv$-prediction, we solve the system of linear equations:
\begin{align}
    \label{eq:system}
    \left\{\begin{aligned}
    \rvz_t &= t \hat{\rvx} + (1 - t) \hat{\rvepsilon} \\
    \rvv_\theta &= \hat{\rvx} - \hat{\rvepsilon}
\end{aligned}\right.
\end{align}
and derive $\hat{\rvx} = \rvz_t + (1 - t) \, \rvv_\theta$ and $\hat{\rvepsilon} = \rvz_t - t \, \rvv_\theta$. Full conversions from any network prediction to any target is shown in \cref{tab:conversions}.

\paragraph{Loss Space.}
\label{par:loss}
The loss is often defined in the space of the network output prediction, \eg \cref{eq:flow-loss} for $\rvv$-prediction. However, the loss target can be defined in any of the three target spaces, and it has been shown that with a given reparameterization from an output prediction to a loss target, the loss is equivalent to a weighted loss in the output prediction space~\cite{karras2022elucidating,salimans2022progressive,esser2024scaling,li2026back}. For example, with $\rvx$-prediction, the loss in the $\rvv$-space can be derived by recovering $\hat{\rvv}$ and $\rvv$ from $\rvx_\theta$ and $\rvx$ using \cref{tab:conversions}:
\begin{align}
    \mathcal{L} &= \mathbb{E} \big\| \hat{\rvv} - \rvv \big\|^2 \nonumber\\
    &= \mathbb{E} \bigg\| \left( \frac{\rvx_\theta - \rvz_t}{1-t} \right) - \left( \frac{\rvx - \rvz_t}{1-t} \right) \bigg\|^2 \quad \text{(via~\cref{tab:conversions})} \nonumber\\
    &= \mathbb{E} \underbrace{\frac{1}{(1-t)^2}}_{\text{weighting over } t} \big\| \rvx_\theta - \rvx \big\|^2.
    \label{eq:loss-weight-example}
\end{align}

\paragraph{Sampling Space.}
\label{par:sampling-space}
Following \citet{li2026back}, regardless of the prediction$\times$loss combination used, we always convert the network output to velocity $\rvv$ to perform sampling by solving the ODE using a numerical solver. See appendix for training and sampling algorithms.

\section{Experimental Setup}
\label{sec:setup}
A diffusion human motion model requires making three possibly-coupled choices: the motion representation, the output of the network, and the space of the loss. We fix the first choice to raw joint positions as they need no hand-engineered features and processing, no autoencoder and lossy reconstruction, and allow for direct spatial control. Unlike a decoder or redundant features that absorb part of the model's error, this representation leaves nothing between the network and the data, allowing us to focus on the effect of the output and loss.

\paragraph{Baseline.}
As a controlled baseline, we adopt the ACMDM~\cite{meng2025absolute} architecture: a Diffusion Transformer (DiT)~\cite{peebles2023scalable} that achieves SOTA T2M performance on AE latents of absolute joint positions, and comparable performance when trained on the raw positions directly. Following the protocol of \citet{li2026back} for raw pixel images, the architecture, solver, and sampling budget stay constant throughout, while the $t$-sampler, model size and motion representations are varied one at a time for additional control. In what follows, we refer to \emph{Baseline} as the recipe of ACMDM~\cite{meng2025absolute} for operating on raw joint positions which uses $\rvv$-prediction and $\rvv$-loss, is trained with uniform $t$ sampling, and samples with an adaptive Dopri5 solver. We use \emph{Ours} to refer to the best recipe we found which uses $\rvx$-prediction and $\rvv$-loss, is trained with logit-normal $t$ sampling, and samples with a fixed Euler solver.

\paragraph{Motion Representation.}
\label{sec:setup-representation}
All models operate on absolute joint positions $\rvx \in \mathbb{R}^{N \times J \times 3}$, with $J=22$ joints for HumanML3D and $J=21$ joints for KIT-ML. For representation control, we also train on the original relative-local redundant HumanML3D representation, which stacks root velocities, root-relative positions, root-relative rotations, global joint velocities and foot contacts into $\rvx \in \mathbb{R}^{N \times 263}$, and on the latents of ACMDM's causal autoencoder, a 2D ResNet downsampling the sequence along the length yeilding the 4-channel latent per joint embedding of $\rvx \in \mathbb{R}^{N/4 \times J \times 4}$. Both are mapped back to joint positions before evaluation.

\paragraph{Protocol.}
\label{sec:setup-protocol}
We train on the HumanML3D~\cite{guo2022generating} dataset and validate our findings on the smaller KIT-ML~\cite{plappert2016kit} dataset, using ACMDM's S model (8 layers, 39M parameters) and training recipe unless stated otherwise. We train HumanML3D models for 500 epochs (191K iterations) and KIT-ML models for 55K iterations. We use a 50-step Euler sampler and classifier guidance scale of 3.0 at inference unless stated otherwise. We report the standard T2M metrics: FID, R-Precision, MM-Dist, Diversity and MultiModality with the evaluator of \citet{meng2025rethinking} on HumanML3D, following ACMDM, and that of \citet{guo2022generating} on KIT-ML. Because these metrics do not fully capture visually-observable motion quality, we add additional metrics that quantify bone-length change, joint acceleration and jerk, and floor penetration, each compared against the real test set motions. Every reported number is the mean over 20 evaluation repeats with its 95\% confidence interval unless stated otherwise. Following \citet{li2026back}, we floor the denominators of the loss weights to stabilize training. At sampling, $\rvepsilon$-prediction models divide by $t$ to convert output to velocity, making them extra sensitive to errors when $t \to 0$. So for these models, we also evaluate a floored sampler and report the better of the two in FID, marked with $\dagger$ (see appendix for full discussion).

\section{What Matters for Diffusion on Raw Motion}
\label{sec:results}
With the representation fixed, we study what the network should predict, in which space the loss should be measured, and how the answer depends on the choices we held fixed. We refer to a model by its output and loss space, \eg $\rvx$-$\rvv$ for $\rvx$-prediction with $\rvv$-loss.

\begin{table*}[t]
  \caption{Prediction target \vs loss space on HumanML3D.
  $\downarrow$: lower is better, $\uparrow$: higher is better, $\rightarrow$: closer to Real is better.
  For $\rvepsilon$-prediction we report the better in FID of exact and floored sampling; $\dagger$ marks floored sampling.
  Bold marks the best of the nine models in each column, except for multimodality.
  The last two rows are the complete recipes: Baseline is $\rvv$-$\rvv$ sampled with Dopri5, Ours is $\rvx$-$\rvv$ with the logit-normal $t$-sampler.}
  \label{tab:hml3d-3x3}
  \centering
  \footnotesize
  \setlength{\tabcolsep}{1.8pt}
  \begin{tabular}{@{}ccccccccccccc@{}}
    \toprule
    & & \multicolumn{7}{c}{Text-to-Motion metrics} & \multicolumn{4}{c}{Physical plausibility} \\
    \cmidrule(lr){3-9} \cmidrule(l){10-13}
    & & & \multicolumn{3}{c}{R-Precision$\uparrow$} & & & & & & & \\
    \cmidrule(lr){4-6}
    Pred. & Loss & FID$\downarrow$ & Top 1 & Top 2 & Top 3 & MM-Dist$\downarrow$ & Diversity$\rightarrow$ & MModality$\uparrow$
      & Bone$\downarrow$ & Accel.$\rightarrow$ & Jerk$\rightarrow$ & Floor$\downarrow$ \\
    & & & & & & & & & mm/frame & m/s$^2$ & m/s$^3$ & mm \\
    \midrule
    \multicolumn{2}{c}{Real}
      & -- & 0.500$^{\pm.002}$ & 0.694$^{\pm.003}$ & 0.793$^{\pm.002}$ & 3.293$^{\pm.008}$ & 10.506$^{\pm.105}$ & --
      & 0.000$^{\pm.000}$ & 2.764$^{\pm.001}$ & 48.9$^{\pm0.0}$ & 0.000$^{\pm.000}$ \\
    \midrule
    $\rvv$ & $\rvv$
      & 0.149$^{\pm.007}$ & 0.499$^{\pm.003}$ & 0.693$^{\pm.003}$ & 0.791$^{\pm.003}$ & 3.295$^{\pm.010}$ & \textbf{10.725}$^{\pm.078}$ & 2.165$^{\pm.076}$
      & 2.712$^{\pm.009}$ & 4.457$^{\pm.011}$ & 120.2$^{\pm0.4}$ & 0.076$^{\pm.004}$ \\
    $\rvv$ & $\rvepsilon$
      & 0.400$^{\pm.015}$ & 0.412$^{\pm.003}$ & 0.596$^{\pm.003}$ & 0.702$^{\pm.003}$ & 3.960$^{\pm.010}$ & 10.962$^{\pm.096}$ & 3.385$^{\pm.124}$
      & 1.469$^{\pm.004}$ & 3.465$^{\pm.009}$ & 79.4$^{\pm0.2}$ & \textbf{0.051}$^{\pm.002}$ \\
    $\rvv$ & $\rvx$
      & 1.662$^{\pm.030}$ & 0.447$^{\pm.003}$ & 0.635$^{\pm.002}$ & 0.739$^{\pm.002}$ & 3.599$^{\pm.011}$ & 9.592$^{\pm.073}$ & 2.165$^{\pm.069}$
      & 10.07$^{\pm.03}$ & 10.69$^{\pm.03}$ & 366.0$^{\pm0.9}$ & 0.137$^{\pm.006}$ \\
    \midrule
    $\rvepsilon$\floored & $\rvv$
      & 1.118$^{\pm.015}$ & 0.457$^{\pm.003}$ & 0.639$^{\pm.003}$ & 0.738$^{\pm.002}$ & 3.571$^{\pm.011}$ & 9.810$^{\pm.067}$ & 2.555$^{\pm.087}$
      & 7.046$^{\pm.018}$ & 8.520$^{\pm.018}$ & 283.9$^{\pm0.7}$ & 0.096$^{\pm.009}$ \\
    $\rvepsilon$\floored & $\rvepsilon$
      & 2.174$^{\pm.033}$ & 0.357$^{\pm.003}$ & 0.530$^{\pm.002}$ & 0.639$^{\pm.002}$ & 4.315$^{\pm.012}$ & 9.667$^{\pm.075}$ & 3.619$^{\pm.132}$
      & 6.836$^{\pm.018}$ & 8.241$^{\pm.020}$ & 272.8$^{\pm0.7}$ & 0.262$^{\pm.018}$ \\
    $\rvepsilon$\floored & $\rvx$
      & 4.910$^{\pm.042}$ & 0.391$^{\pm.003}$ & 0.567$^{\pm.003}$ & 0.671$^{\pm.003}$ & 4.106$^{\pm.010}$ & 8.687$^{\pm.061}$ & 2.415$^{\pm.103}$
      & 12.17$^{\pm.03}$ & 12.84$^{\pm.04}$ & 450.5$^{\pm1.3}$ & 0.148$^{\pm.005}$ \\
    \midrule
    $\rvx$ & $\rvv$
      & \textbf{0.134}$^{\pm.004}$ & \textbf{0.508}$^{\pm.003}$ & \textbf{0.700}$^{\pm.003}$ & \textbf{0.797}$^{\pm.003}$ & \textbf{3.251}$^{\pm.010}$ & 10.803$^{\pm.075}$ & 2.057$^{\pm.081}$
      & 1.573$^{\pm.003}$ & 3.646$^{\pm.010}$ & 83.0$^{\pm0.2}$ & 0.055$^{\pm.003}$ \\
    $\rvx$ & $\rvepsilon$
      & 0.304$^{\pm.015}$ & 0.396$^{\pm.003}$ & 0.580$^{\pm.002}$ & 0.687$^{\pm.002}$ & 4.046$^{\pm.010}$ & 10.869$^{\pm.089}$ & 3.433$^{\pm.119}$
      & \textbf{1.409}$^{\pm.002}$ & 3.262$^{\pm.008}$ & 74.7$^{\pm0.1}$ & 0.077$^{\pm.008}$ \\
    $\rvx$ & $\rvx$
      & 0.197$^{\pm.006}$ & 0.490$^{\pm.003}$ & 0.683$^{\pm.003}$ & 0.782$^{\pm.002}$ & 3.369$^{\pm.009}$ & 10.779$^{\pm.083}$ & 2.101$^{\pm.074}$
      & 1.851$^{\pm.004}$ & \textbf{3.150}$^{\pm.009}$ & \textbf{62.6}$^{\pm0.2}$ & 0.069$^{\pm.002}$ \\
    \bottomrule
  \end{tabular}
\end{table*}

\subsection{Prediction Output $\times$ Loss Target}
\label{sec:results-grid}

\begin{figure}[t]
  \centering
  \includegraphics[width=0.87\linewidth]{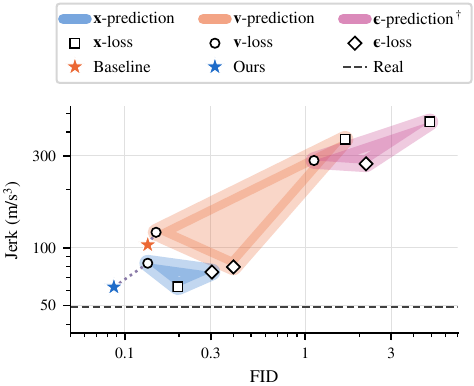}
  \caption{All nine prediction$\times$loss combinations on HumanML3D, with FID on the horizontal axis and Jerk on the vertical axis (both on logarithmic scale); lower left is better. $\rvepsilon$-prediction models are sampled with the floored conversion ($\dagger$).
  The stars are the complete recipes: Baseline is ACMDM~\cite{meng2025absolute} ($\rvv$-$\rvv$ sampled with Dopri5) and Ours is $\rvx$-$\rvv$ with the logit-normal $t$-sampler; a dotted line links each to the corresponding model it builds on.
  The dashed line is the jerk of the real test motions.}
  \label{fig:hml3d-fid-jerk}
\end{figure}

\paragraph{$\rvx$-prediction performs best.}
In \cref{tab:hml3d-3x3}, every $\rvx$-prediction model has good performance, whichever loss it is train with. Every $\rvepsilon$-prediction model has poor performance, whichever loss it is trained with. $\rvv$-prediction sits in between and is sensitive to the loss: it is competitive with $\rvv$-loss and does poorly with $\rvx$-loss. Because sampling converts every output to velocity, $\rvepsilon$-prediction divides by $t$ and is very sensitive to errors near $t=0$ (\cref{tab:conversions}). Placing a floor on the denominator helps some $\rvepsilon$-prediction models and hurt others, so in all tables we report the better of the two samplers in FID, marking the floored results with $\dagger$ (see appendix for details). Even with this adjustment, $\rvepsilon$-prediction performs the worst. \Cref{fig:hml3d-fid-jerk} shows the same ordering in the FID-Jerk plane: the three $\rvx$-prediction models lie closest to the bottom left corner, $\rvv$-prediction models above them, and the $\rvepsilon$-prediction models far from the other two.

\paragraph{The choice of loss is of secondary importance.}
Changing the loss to another space is equivalent to reweighting it over $t$ (\cref{sec:prediction-loss}). The results of \cref{tab:hml3d-3x3} show that no weighting makes an $\rvepsilon$-prediction model produce high quality and smooth motions, while a poor choice can strongly degrade the performance of a $\rvv$-prediction model. For $\rvx$-prediction models, the choice of loss is largely a matter of preference: $\rvv$-loss has better fidelity and text alignment, and $\rvx$-loss has better smoothness. In \cref{fig:hml3d-fid-jerk}, the size of each region reflects this sensitivity: the choice of loss has only a small impact on an $\rvx$-prediction model, while far more on a $\rvv$-prediction model.
We use $\rvx$-$\rvv$ as our preferred model for the remainder of our results.

\paragraph{Standard T2M quality metrics hide the difference.}
Judging by FID, R-Precision and Diversity, $\rvv$-$\rvv$ and $\rvx$-$\rvv$ models are comparable (\cref{tab:hml3d-3x3}), which is why velocity-prediction on raw joints is sometimes seen as a solved problem~\cite{meng2025absolute}. However, the quality of the visualized motions, provided as supplementary video, and as captured by the plausibility metrics (\cref{tab:hml3d-3x3}) reveals a different result: $\rvv$-$\rvv$ produces jittery motions with higher Acceleration, Jerk and Bone-length Change compared to $\rvx$-$\rvv$. The standard T2M metrics employ a motion embedding, thereby smoothing out high-frequency errors, making them oblivious to motion jitter. We report a number of physical plausibility metrics, including the motion Jerk (the time derivative of acceleration), in order to better capture the motion quality as visible to a viewer.

\begin{figure*}[t]
  \centering
  \includegraphics[width=0.85\linewidth]{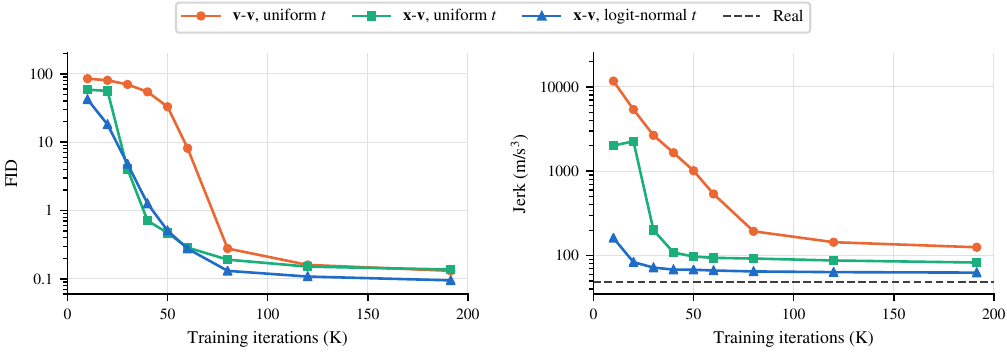}
  \caption{Training curves of our models on HumanML3D (raw joint positions). FID and Jerk are on logarithmic scale. Each point is a checkpoint evaluated with 5 repeats.}
  \label{fig:hml3d-curves}
\end{figure*}

\paragraph{$\rvx$-prediction is more training-efficient.}
\label{sec:res-x-pred-training-efficient}
Although we use the same training recipe across all prediction-loss combinations, $\rvx$-$\rvv$ reaches an FID below 1 by 40K iterations, while $\rvv$-$\rvv$ requires 80K iterations (\cref{fig:hml3d-curves}). Replacing the uniform $t$-sampler of $\rvx$-$\rvv$ with a logit-normal sampler~\cite{esser2024scaling} improves it even further, as \citet{li2026back} observed for images.  The logit-normal sampler draws $\xi \sim \mathcal{N}(\mu, \sigma^2)$ and sets $t = \operatorname{sigmoid}(\xi)$. We set $\mu=-0.8$ and $\sigma=0.8$ following \citet{li2026back}, which concentrates training on noisier inputs (see appendix for discussion and ablations on choice of $\mu$). $\rvx$-$\rvv$ with the logit-normal $t$-sampler matches the final FID of $\rvv$-$\rvv$ with half its jerk, using only 42\% of the 191.5K training budget (\cref{fig:hml3d-curves}). We observe that the same logit-normal $t$-sampler hurts the performance of $\rvv$-$\rvv$ (see appendix). We pick $\rvx$-$\rvv$ with logit-normal $t$ sampling as the top recipe in this work (referred to as \emph{Ours}).

\begin{table}[t]
  \caption{Sampling cost of Dopri5 on $\rvv$-$\rvv$ and Euler on $\rvx$-$\rvv$, as in \cref{tab:hml3d-recipes-compact}. Solvers are timed on one NVIDIA H100 GPU, at batch size 1 (latency of a single prompt) and batch size 32 (bulk generation).
  NFE is the number of network evaluations per sample.
  We report the mean and standard deviation over 50 timed batches.}
  \label{tab:hml3d-cost}
  \centering
  \footnotesize
  \begin{tabular}{@{}lcccc@{}}
    \toprule
    & \multicolumn{2}{c}{Batch size 1} & \multicolumn{2}{c}{Batch size 32} \\
    \cmidrule(lr){2-3} \cmidrule(l){4-5}
    Solver & NFE & ms/sample & NFE & ms/sample \\
    \midrule
    Euler, 50 steps & \textbf{50} & \textbf{357}$\pm$4 & \textbf{50} & \textbf{31.7}$\pm$0.0 \\
    Dopri5 & 149$\pm$26 & 1072$\pm$184 & 208$\pm$23 & 132.0$\pm$14.6 \\
    \bottomrule
  \end{tabular}
\end{table}

\begin{table}[t]
  \caption{Top recipes: top performers of our 3$\times$3 prediction-time analysis of \cref{tab:hml3d-3x3} ($\rvx$-$\rvv$ and $\rvv$-$\rvv$) using a 50-step Euler solver \vs the recipe of ACMDM~\cite{meng2025absolute} ($\rvv$-$\rvv$, Dopri5 solver) \vs our recommended recipe on HumanML3D ($\rvx$-$\rvv$, logit-normal $t$-sampler, 50-step Euler solver). Values are means over 20 evaluation repeats (confidence intervals and the remaining metrics in appendix).
  $\downarrow$: lower is better, $\uparrow$: higher is better, $\rightarrow$: closer to the real test motions is better (Diversity 10.506, Acceleration 2.764\,m/s$^2$, Jerk 48.9\,m/s$^3$).
  Bold marks the best model in each column.}
  \label{tab:hml3d-recipes-compact}
  \centering
  \footnotesize
  \setlength{\tabcolsep}{3.6pt}
  \begin{tabular}{@{}lccccc@{}}
    \toprule
    & \multicolumn{3}{c}{Text-to-Motion metrics} & \multicolumn{2}{c}{Physical plausibility} \\
    \cmidrule(lr){2-4} \cmidrule(l){5-6}
    Model & FID$\downarrow$ & Top 1$\uparrow$ & Diversity$\rightarrow$ & Accel.$\rightarrow$ & Jerk$\rightarrow$ \\
    & & & & m/s$^2$ & m/s$^3$ \\
    \midrule
    $\rvv$-$\rvv$ & 0.149 & 0.499 & 10.725 & 4.457 & 120.2 \\
    $\rvx$-$\rvv$ & 0.134 & \textbf{0.508} & 10.803 & 3.646 & 83.0 \\
    \midrule
    Baseline & 0.134 & 0.497 & 10.720 & 4.160 & 103.4 \\
    Ours  & \textbf{0.087} & 0.475 & \textbf{10.507} & \textbf{3.070} & \textbf{62.4} \\
    \bottomrule
  \end{tabular}
\end{table}

\paragraph{A simple solver is enough.}
ACMDM samples the $\rvv$-$\rvv$ model with the adaptive Dopri5 solver (referred to as \emph{Baseline}), which improves over Euler, but takes three to four times as many network evaluations (\cref{tab:hml3d-cost}). $\rvx$-$\rvv$ achieves better performance with a simple 50-step Euler solver, matching the FID of $\rvv$-$\rvv$ with Dopri5 and has lower Jerk. With the logit-normal $t$-sampler, $\rvx$-$\rvv$ is better in FID, Acceleration, and Jerk while using the same fixed Euler sampler. Performance of these top recipes are provided in \cref{tab:hml3d-recipes-compact}.

\begin{table}[t]
  \caption{Prediction target \vs loss space on KIT-ML; each cell gives FID$\downarrow$ / Jerk$\rightarrow$ ($10^3$\,u/s$^3$, in the native units of KIT-ML).
  The Jerk of the real test motions is 37.3.
  Bold marks the best FID and the Jerk closest to the real value.
  $\dagger$ marks floored sampling.}
  \label{tab:kit-grid-combined}
  \centering
  \footnotesize
  \renewcommand{\arraystretch}{1.15}
  \begin{tabular}{@{}l*{3}{>{\centering\arraybackslash}p{0.23\linewidth}}@{}}
    \toprule
    & $\rvv$-loss & $\rvepsilon$-loss & $\rvx$-loss \\
    \midrule
    $\rvv$-pred & 1.139 / 122.3 & 0.511 / 93.8 & 26.32 / 442.3 \\
    $\rvepsilon$-pred & 0.782 / 176.6 & 0.505 / 64.3\floored & 2.440 / 166.8 \\
    $\rvx$-pred & \textbf{0.176} / \textbf{39.2} & 0.203 / 39.9 & 0.196 / 31.7 \\
    \bottomrule
  \end{tabular}
\end{table}

\paragraph{Results hold across datasets.}
KIT-ML experiments reproduce the same trends as HumanML3D (\cref{tab:kit-grid-combined}). With a 55K-iteration budget, $\rvx$-prediction models have all converged, and $\rvx$-$\rvv$ has the best FID and the Jerk closest to real motion among the nine combinations; it is also best in R-Precision and MM-Dist (see appendix). $\rvv$-$\rvv$ converges more slowly: it needs the full 191.5K iterations to reach a comparable FID, and its Jerk remains higher than that of $\rvx$-$\rvv$ at 55K iterations (see full evaluation table and training curves in appendix).

\subsection{Understanding when and why \texorpdfstring{$\rvx$}{x}-\texorpdfstring{$\rvv$}{v} models improve on \texorpdfstring{$\rvv$}{v}-\texorpdfstring{$\rvv$}{v} models}
\label{sec:results-when}
In prior section, we fix the model size and train on raw joint positions and show $\rvx$-$\rvv$ is the strongest combination. We now change the model capacity, the width of the input embedding, and the representation, to understand when the gap between $\rvx$-prediction and $\rvv$-prediction holds.

\begin{table}[t]
  \caption{FID$\downarrow$ / Jerk$\rightarrow$ (m/s$^3$) of the final models (191.5K iterations) at each model size on HumanML3D.
  We report the mean over 20 evaluation repeats; the Jerk of the real test motions is 48.9\,m/s$^3$.
  Bold marks the best FID and the Jerk closest to the real motions.}
  \label{tab:hml3d-size-fid-jerk}
  \centering
  \footnotesize
  \setlength{\tabcolsep}{5pt}
  \begin{tabular}{@{}lccc@{}}
    \toprule
    Model & XS & S & L \\
    \midrule
    $\rvv$-$\rvv$, uniform $t$ & 0.349 / 161.7 & 0.149 / 120.2 & 0.116 / 108.5 \\
    $\rvx$-$\rvv$, uniform $t$ & 0.214 / 98.2 & 0.134 / 83.0 & 0.103 / 75.7 \\
    $\rvx$-$\rvv$, logit-normal $t$ & 0.178 / 65.6 & \textbf{0.087} / 62.4 & 0.166 / \textbf{60.8} \\
    \bottomrule
  \end{tabular}
\end{table}

\paragraph{Capacity does not improve $\rvv$-prediction jitter.}
If the jitter of $\rvv$-$\rvv$ is due to an undersized model, a larger model would help. Growing the model size from 5M (XS, 4 layers) to 130M (L, 12 layers) parameters lowers the Jerk of $\rvv$-$\rvv$ by a third, but still underperforms the 5M $\rvx$-$\rvv$ model in terms of measured Jerk, both with and without the logit-normal $t$-sampler (\cref{tab:hml3d-size-fid-jerk}). In contrast, $\rvx$-prediction models see only marginal improvements to motion jitter (as measured by Jerk) when larger models are used (\cref{tab:hml3d-size-fid-jerk}). See appendix for training curves on FID and Jerk.

\begin{figure}[t]
  \centering
  \includegraphics[width=0.85\linewidth]{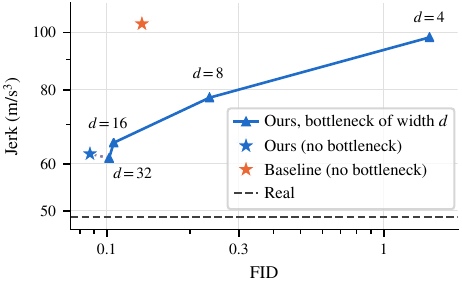}
  \caption{Effect of a low-rank bottleneck of width $d$ in the frame embedding on HumanML3D.
  The points are our $\rvx$-$\rvv$ model with the logit-normal $t$-sampler; the stars are the models without a bottleneck.
  The dashed line is the jerk of the real test motions. Both axes are on a logarithmic scale.}
  \label{fig:hml3d-bottleneck-fid-jerk}
\end{figure}

\paragraph{$\rvx$-prediction tolerates a bottleneck, $\rvv$-prediction does not.}
Following \citet{li2026back}, we replace the linear embedding of each frame with a low-rank one of width $d \in {4, 8, 16, 32}$. As shown in \cref{fig:hml3d-bottleneck-fid-jerk}, $\rvx$-prediction (in blue) tolerates the bottleneck, i.e.  compressing $22 \times 3$-dimensional frames to 16 dimensions costs little in terms of FID and Jerk and at $d=8$ the motion is smoother than that of the full-rank $\rvv$-$\rvv$ model (in orange). In contrast, $\rvv$-$\rvv$ fails to train for any of the reduced-rank bottlenecks. The network sees $\rvz_t$ only through the bottleneck, so the noise outside the $d$ dimensions is invisible to it. A $\rvv$-predictor must predict the noise as part of $\rvv = \rvx - \rvepsilon$, and it cannot (see full evaluation results in appendix). An $\rvx$-predictor returns the motion and because 16 dimensions is sufficient, we can conclude that the poses and motions do lie on a lower-dimensional manifold and that $\rvx$-prediction models can make use of this. Unlike in the case of images~\cite{li2026back}, the bottleneck does not improve FID and thus we keep the full-rank linear embedding.

\paragraph{Other representations close the gap, at a cost.}
The gap between $\rvx$ and $\rvv$ prediction is special to raw joints (\cref{tab:hml3d-repr-fid-jerk}). On autoencoder (AE) latents, the gap disappears: $\rvv$-$\rvv$ and $\rvx$-$\rvv$ reach nearly the same FID and Jerk, with $\rvv$-$\rvv$ slightly ahead. This was expected as the AE already compresses the motion, so the network operates on low-dimensional space, regardless of its output prediction, and the decoder smooths the output to produce the smooth joint-position predictions. Redundant features also remove the jitter of $\rvv$-prediction, as expected of a structured representation with a plausibility prior built into the data. That prior costs fidelity: in every column, the redundant model has worse FID than its raw-joint variant. The logit-normal $t$-sampler also lowers FID only on raw joints. This sampler and $\rvx$-prediction on raw joints match the latent-model results in Jerk and reach the best FID on the table, with neither an AE nor hand-designed features. See appendinx for full evaluation results on both representations.

\subsection{A Simple Recipe for Raw Motion}
\label{sec:results-recipe}
When operating on raw joint positions, our findings suggest the following simple recipe: predict $\rvx$, compute the loss on $\rvv$, sample $t$ from a logit-normal distribution at training, and use a simple-and-efficient fixed Euler sampler. Predicting $\rvx$ is what removes the jitter, on both datasets and at every model size, and neither the loss, nor capacity, nor an adaptive solver recovers it for $\rvv$-prediction. $\rvx$-prediction matters most when training budget is limited, the model is small, or it passes through a bottleneck. The other three choices build on it: the $\rvv$-loss gives $\rvx$-prediction the best fidelity, the logit-normal sampler lowers FID and Jerk further (at a slightly lower R-Precision, see \cref{tab:hml3d-3x3}), and a fixed-step solver is enough once the output is $\rvx$.

\section{Conclusions and Limitations}
\label{sec:conclusions}
Raw joint positions are a simple representation for diffusion-based human motion models as they need no conversions and directly support spatial control. However, with standard choices of prediction target and loss objective, generated motions show visible jitter that commonly reported text-to-motion metrics such as FID fail to capture. We study the prediction$\times$loss space and find a simple and effective recipe for smooth raw motion generation. Predict the clean motion ($\rvx$-prediction), use velocity as the loss objective ($\rvv$-loss), sample $t$ form a logit-normal distribution during training, and use a simple Euler solver at inference. This recipe outperforms other model combinations on a variety of metrics, including fidelity metrics such as FID and plausibility metrics designed to measure motion jitter such as Jerk. We observe that the relative benefit of $\rvx$-prediction is particularly significant, which we hypothesize is due to the low intrinsic dimensionality of human motion data. Furthermore, $\rvx$-prediction speeds training convergence, specially for low-capacity models, suggesting a less expensive starting point for many possible motion tasks. We further envision that reporting plausibility metrics that capture motion jitter becomes common practice.

Our work comes with a number of limitations. For our controlled study, we fix several design choices, and these define (and thus possibly limit) the scope of our work. We use a single Diffusion Transformer backbone and flow matching formulation with a linear schedule. We vary the backbone over three sizes, from 5M to 130M parameters. There exist further possible architecture choices, noise schedules, and diffusion paradigms which we leave to future work. To measure plausibility, we rely on kinematic statistics which agree with our observations in generated motions and are easy to compute and report, but are only proxies for perceptual visual quality. 
Testing our proposed diffusion design on other motion tasks such as motion editing is a natural next step.

\begin{table}[t]
  \caption{FID$\downarrow$ / Jerk$\rightarrow$ (m/s$^3$) of each model on the three motion representations of HumanML3D. Jerk is measured on the joint positions recovered from each representation.
  We report the mean over 20 evaluation repeats; the Jerk of the real test motions is 48.9\,m/s$^3$.
  Bold marks, for each representation, the best FID and the Jerk closest to the real motions.}
  \label{tab:hml3d-repr-fid-jerk}
  \centering
  \footnotesize
  \setlength{\tabcolsep}{4pt}
  \begin{tabular}{@{}lccc@{}}
    \toprule
    & $\rvv$-$\rvv$ & $\rvx$-$\rvv$ & $\rvx$-$\rvv$ \\
    Representation & uniform $t$ & uniform $t$ & logit-normal $t$ \\
    \midrule
    Raw joints & 0.149 / 120.2 & 0.134 / 83.0 & \textbf{0.087} / \textbf{62.4} \\
    \midrule
    AE latents & \textbf{0.142} / \textbf{62.7} & 0.172 / 66.2 & 0.571 / 89.9 \\
    \midrule
    Redundant features & \textbf{0.399} / 52.3 & 0.513 / 81.1 & 0.736 / \textbf{48.2} \\
    \bottomrule
  \end{tabular}
\end{table}

\section*{LLM Usage Statement}
We used generative AI tools (Anthropic's Claude models and Google Gemini) for coding and writing assistance under author supervision. On the coding side, this covered bookkeeping of experimental results, experiment-specific code modifications, plotting and rendering scripts, and \LaTeX{} formatting of figures and tables including filling tables from evaluation logs. On the writing side, it covered organizing the presentation of results, improving the clarity of some passages, grammar correction, and help with finding relevant references. The authors reviewed all LLM-assisted work, traced reported numbers to their evaluation logs, verified all cited references, and take full responsibility for the content of this paper.

{
    \small
    \bibliographystyle{ieeenat_fullname}
    \bibliography{main}
}

\appendix
\clearpage
\maketitlesupplementary

\noindent This appendix is organized as follow. \Cref{supp:unified-parameterization} defines diffusion and flow matching under a single parameterization. \Cref{supp:loss-weights,sec:algorithms} detail the loss weights used in our implementation of network output-to-loss conversions and the exact training and sampling algorithms used in this work. \Cref{supp:flooring} provides details regarding  the flooring of the conversion denominators $t$ and $1-t$, which stabilizes training but must be used with care at sampling time. \Cref{sec:implementation-details} contains details on datasets, architecture, training configuration and evaluation metrics. \Cref{supp:evaluator} provides a discussion on the choice of evaluator. \Cref{sec:supp-tsampler} discusses the logit-normal $t$-sampler, its impacts on $\rvv$-prediction and $\rvx$-prediction, and provides a sweep over the hyperparameter $\mu$. Finally, \cref{sec:additional-results} provides the complete tables and additional plots of what is reported in a compact form in the main body of the paper. These include the best-choice recipes, KIT-ML, model sizes, bottlenecks, other motion representations, and flooring at sampling. 

\noindent
Videos of generated motions are provided as separate supplementary material.

\section{Unified Parameterization}
\label{supp:unified-parameterization}
Diffusion and flow matching models are trained on noisy samples $\rvz_t = \alpha_t \rvx + \sigma_t \rvepsilon$, where $\rvx$ is the clean data sample, $\rvepsilon \sim \mathcal{N}(\mathbf{0}, \mathbf{I})$ is Gaussian noise, and the noise schedule $\alpha_t$ and $\sigma_t$ distinguishes DDPM, score-based, and flow-matching paradigms (\cref{tab:parameterization-schedules}). The network is trained to predict a target $\rvu_t = \rho_t \rvx + \phi_t \rvepsilon$, which can be used for each of noise, clean-sample, velocity, and score prediction for different choices of $\rho_t$ and $\phi_t$ (\cref{tab:parameterization-targets}), by minimizing a loss function defined as:
\begin{align}
  \mathcal{L}(\theta)
  = \mathbb{E}_{t, \rvx, \rvepsilon}
  \Big[ w_{\rvu \to \boldsymbol{\ell}}(t) \, \big\| \rvu_\theta(\rvz_t, t) - \rvu_t \big\|^2 \Big],
  \label{eq:general-loss}
\end{align}
where $w_{\rvu \to \boldsymbol{\ell}}(t)$ is a weighting function converting the loss space from $\rvu$ to $\boldsymbol{\ell}$. Since $\rvz_t$ and $\rvu_t$ are both linear combinations of $(\rvx, \rvepsilon)$, any target can be mapped to any other given $\rvz_t$ by a linear map when $\alpha_t \phi_t - \sigma_t \rho_t \neq 0$, which holds for all targets above at every diffusion timestep $t$. Therefore, the paradigms are the same model but with different design choices for the noise schedule, the loss weighting over $t$, and the sampler that enables generation from the trained network. This allows us to (i) treat the prediction target and the loss space as independent choices, and (ii) study them within a single paradigm without loss of generality.

\section{Loss weights}
\label{supp:loss-weights}
Under the flow matching paradigm with a linear schedule, to change the loss space from the network output to some loss target, we multiply the output loss by the $t$-dependent loss weight ($w_{\rvu \to \boldsymbol{\ell}}(t)$ in \cref{alg:training}), corresponding to the corresponding output$\times$loss pair shown in \cref{tab:loss-weights}.

\section{Training and Sampling Algorithms}
\label{sec:algorithms}

\paragraph{Training.} 
The training and sampling algorithms are given in \cref{alg:training,alg:sampling}. The configuration is defined by the network prediction output $\rvu \in \{\rvv, \rvepsilon, \rvx\}$, the loss space $\boldsymbol{\ell} \in \{\rvv, \rvepsilon, \rvx\}$ and the sampler $p(t)$ over diffusion steps, which is either uniform on $[0,1]$ or logit-normal with $t = \operatorname{sigmoid}(\xi)$ where $\xi \sim \mathcal{N}(\mu, \sigma^2)$. Training minimizes the loss in the output space scaled by the weight $w_{\rvu \to \boldsymbol{\ell}}(t)$ set according to \cref{tab:loss-weights}. During training, we follow \citet{li2026back} and floor the conversion denominator ($t$ and $1-t$) to $\delta=0.05$ to keep loss weights bounded and stabilize training (see \cref{supp:flooring}). Text conditioning is dropped with probability $p_\text{uncond} = 0.1$ during training, to enable classifier-free guidance~\cite{ho2022classifier} at sampling time. 

\begin{table}[t]
  \centering
  \small
  \setlength{\tabcolsep}{5pt}
  \caption{Diffusion paradigms and prediction targets in one parameterization. The flow-matching
  velocity is written for the linear schedule $\alpha_t = t$, $\sigma_t = 1-t$ used in this paper.}
  \label{tab:parameterization}
  \captionsetup[subtable]{skip=3pt}
  \begin{subtable}{\linewidth}
    \centering
    \caption{Noise schedules}
    \label{tab:parameterization-schedules}
    \begin{tabular}{@{}lcc@{}}
      \toprule
      Paradigm & $\alpha_t$ & $\sigma_t$ \\
      \midrule
      DDPM~\cite{ho2020denoising} & $\sqrt{\bar\alpha_t}$ & $\sqrt{1-\bar\alpha_t}$ \\
      Score-based~\cite{song2020score} & $1$ & $\sigma_t$ \\
      Flow matching~\cite{lipman2022flow, liu2022flow} & $t$ & $1-t$ \\
      \bottomrule
    \end{tabular}
  \end{subtable}

  \vspace{0.5em}
  \begin{subtable}{\linewidth}
    \centering
    \caption{Prediction targets}
    \label{tab:parameterization-targets}
    \begin{tabular}{@{}lcc@{}}
      \toprule
      Target & $\rho_t$ & $\phi_t$ \\
      \midrule
      Noise $\rvepsilon$ & $0$ & $1$ \\
      Clean sample $\rvx$ & $1$ & $0$ \\
      Flow-matching velocity $\rvx - \rvepsilon$ & $1$ & $-1$ \\
      $\rvv$-prediction~\cite{salimans2022progressive} & $-\sigma_t$ & $\alpha_t$ \\
      Score $\nabla\log p_t(\rvz_t)$ & $0$ & $-1/\sigma_t$ \\
      \bottomrule
    \end{tabular}
  \end{subtable}
\end{table}

\paragraph{Sampling.} 
At sampling time, by default, we use a fixed Euler solver with $N=50$ steps to integrate from noise at $t_0$ to clean motion at $t_1$. At every step $t$, we convert the network output to a velocity estimate according to \cref{tab:conversions} of the paper. We set $[t_0, t_1] = [0, 1]$ with $\rvv$-prediction and $[t_0, t_1] = [10^{-3}, 1-10^{-3}]$ otherwise, where the conversions are finite. Because converting $\rvepsilon$ to $\rvv$ divides by $t$, $\rvepsilon$-prediction models are sensitive to errors near $t = 0$. For these models, we also evaluate a floored sampler that integrates over $[0, 1]$ and floors $t$ at $\delta = 0.05$ in the conversion, as in training. For each model we report the sampler with the better FID, marking the floored one with $\dagger$ (see \cref{supp:flooring}). We apply classifier-free guidance with scale $s = 3$ to the network output before conversion.

\section{Denominator Flooring}
\label{supp:flooring}
Some conversions between network outputs (\cref{tab:conversions} of the paper) divide by $t$ and $1-t$, which converge to 0 at the both ends of the diffusion path. This affects training through the loss weights $w_{\rvu \to \boldsymbol{\ell}}$, and sampling through conversions to velocity. For $\rvx$- and $\rvepsilon$-prediction, $t$ is drawn from $[10^{-3}, 1-10^{-3}]$. $\rvv$-prediction needs no conversion and uses $[0,1]$.

\begin{figure*}[t]
  \centering
  \includegraphics[width=0.95\linewidth]{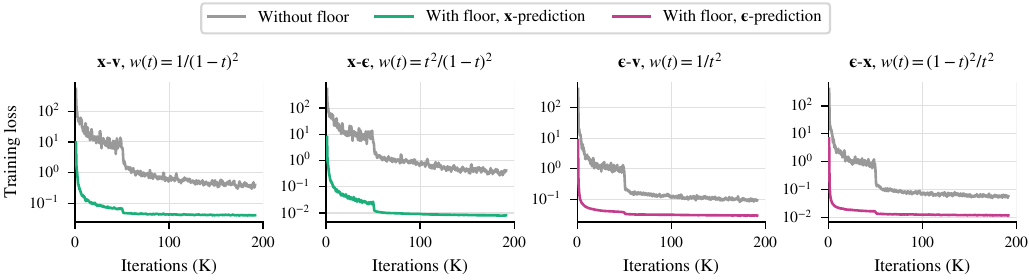}
  \caption{Effect of flooring the denominator of loss weights on training.}
  \label{fig:training-curves-floor}
\end{figure*}

\paragraph{Flooring at Training.}
The loss weights of $\rvepsilon$-prediction and $\rvx$-prediction have denominators as shown in \cref{tab:loss-weights}. For example, with $\rvx$-prediction with $\rvv$-loss, $w(t) = 1/(1-t)^2$ reaches $10^6$ at $t=1-10^{-3}$. This means a few samples near the end of the path dominate the loss and its gradients. Following \citet{li2026back}, we floor the denominators $t$ and $1-t$ at $\delta=0.05$, which bounds the weights by $1/\delta^2 = 400$, keeping the gradients balanced and stabilizing training. \Cref{fig:training-curves-floor} shows the evaluation of training loss with and without flooring for the four cases with such denominators, highlighting the effectiveness of flooring on convergence.

\paragraph{Flooring at Sampling with $\rvx$-prediction.}
\citet{li2026back} clip the denominator ``whenever computing this division'', which in their released codebase includes the conversion of $\rvx$ to $\rvv$ at sampling. We do not follow this. With the floor, one Euler step of size $h$ is:
\begin{align}
    \rvz_{t+h} = \Big(1 - \frac{h}{\max(1-t, \delta)}\Big)\, \rvz_t + \frac{h}{\max(1-t, \delta)}\, \rvx_\theta ,
\end{align}
which means for $1-t < \delta$, flooring lowers the weight of the network output $\rvx_\theta$ and increases the weight of the noisy sample $\rvz_t$. This causes some residual noise to remain for the last steps of even a perfect $\rvx$-predictor. In our evaluation, flooring with $\delta=0.05$ raises the FID of $\rvx$-$\rvv$ from 0.134 to 12.2 and its Jerk from 83.0 to 576\,m/s$^3$ (\cref{tab:hml3d-sampling-floor}). Full results are in \cref{tab:hml3d-sampling-floor-full}.

\begin{table}[t]
  \centering
  \small
  \setlength{\tabcolsep}{2pt}
  \renewcommand{\arraystretch}{1.45}
  \caption{Loss weights. A loss on the column target equals the loss in the network's output
  space (rows) multiplied by the weight in the cell; tinted cells are the unweighted losses.}
  \label{tab:loss-weights}
  \newcolumntype{M}{>{\centering\arraybackslash}p{0.22\linewidth}}
  \begin{tabular}{@{}r@{\hspace{4pt}}MMM@{}}
    \toprule
     & $\|\hat{\rvv} - \rvv\|^2=$ & $\|\hat{\rvepsilon} - \rvepsilon\|^2=$ & $\|\hat{\rvx} - \rvx\|^2=$ \\
    \midrule
    $\|\rvv_\theta - \rvv\|^2 \,\times$ & \cellcolor{gray!14}$1$ & $t^2$ & $(1-t)^2$ \\
    $\|\rvepsilon_\theta - \rvepsilon\|^2 \,\times$ & $1/t^2$ & \cellcolor{gray!14}$1$ & $(1-t)^2/t^2$ \\
    $\|\rvx_\theta - \rvx\|^2 \,\times$ & $1/(1-t)^2$ & $t^2/(1-t)^2$ & \cellcolor{gray!14}$1$ \\
    \bottomrule
  \end{tabular}
\end{table}

\begin{table}[t]
  \caption{Effect of flooring the conversion to velocity when sampling $\rvx$-prediction models, on FID$\downarrow$ and Jerk$\rightarrow$. Each cell gives the metric with the exact conversion $\to$ with the floored one. Model is trained on HumanML3D and real test motions have a Jerk of 48.9 m/s$^3$. Values are means over 20 evaluation repeats. Bold marks the better of the two.}
  \label{tab:hml3d-sampling-floor}
  \centering
  \footnotesize
  \setlength{\tabcolsep}{4pt}
  \begin{tabular}{@{}lcc@{}}
    \toprule
    & FID$\downarrow$ & Jerk$\rightarrow$ (m/s$^3$). \\
    \midrule
    $\rvx$-$\rvv$ & \textbf{0.134}$\to$12.198 & \textbf{83.0}$\to$576.3 \\
    $\rvx$-$\rvv$,logit-norm & \textbf{0.087}$\to$15.069 & \textbf{62.4}$\to$563.4\\
    \bottomrule
  \end{tabular}
\end{table}

\paragraph{Flooring at Sampling with $\rvepsilon$-prediction.}
Converting $\rvepsilon_\theta$ to velocity divides by $t$, so errors of $\rvepsilon_\theta$ are amplified at the start of sampling, near $t = 0$. Flooring $t$ at $\delta$ and integrating over $[0, 1]$ helps control this amplification. It lowers the FID of every model on the raw joints of HumanML3D, but not of all models on the other representation or on KIT-ML (\cref{tab:floor-fid} and \cref{tab:floor-jerk}). We therefore evaluate every $\rvepsilon$-prediction model with both samplers and report the one with the better FID, marking the floored with $\dagger$.

\begin{table}[t]
  \caption{Effect of flooring the conversion to velocity when sampling $\rvepsilon$-prediction models, in FID$\downarrow$.
  Each cell gives the FID with the exact conversion $\to$ with its denominator $t$ floored at $\delta=0.05$.
  Values are means over 20 evaluation repeats.
  Bold marks the better of the two, which is the result reported for that model in all other tables.}
  \label{tab:floor-fid}
  \centering
  \footnotesize
  \setlength{\tabcolsep}{4pt}
  \begin{tabular}{@{}lccc@{}}
    \toprule
    & $\rvepsilon$-$\rvv$ & $\rvepsilon$-$\rvepsilon$ & $\rvepsilon$-$\rvx$ \\
    \midrule
    Raw joints & 1.795\,$\to$\,\textbf{1.118} & 30.60\,$\to$\,\textbf{2.174} & 7.267\,$\to$\,\textbf{4.910} \\
    Redundant 263-dim & 1.476\,$\to$\,\textbf{1.230} & 50.33\,$\to$\,\textbf{6.395} & \textbf{0.753}\,$\to$\,1.450 \\
    AE latents & 0.642\,$\to$\,\textbf{0.082} & 24.90\,$\to$\,\textbf{0.204} & 1.869\,$\to$\,\textbf{1.191} \\
    \midrule
    KIT-ML & \textbf{0.782}\,$\to$\,10.30 & 41.17\,$\to$\,\textbf{0.505} & \textbf{2.440}\,$\to$\,31.52 \\
    \bottomrule
  \end{tabular}
\end{table}

\begin{table}[t]
  \caption{Effect of flooring the conversion to velocity when sampling $\rvepsilon$-prediction models, in Jerk$\rightarrow$.
  Same models, samplers and layout as \cref{tab:floor-fid}: each cell gives the Jerk with the exact conversion $\to$ with the floored one.
  HumanML3D rows are in m/s$^3$ (real test motions: 48.9) and KIT-ML is in thousands of its native units (real test motions: 37.3).
  Bold marks the value closer to the real motions.}
  \label{tab:floor-jerk}
  \centering
  \footnotesize
  \setlength{\tabcolsep}{4pt}
  \begin{tabular}{@{}lccc@{}}
    \toprule
    & $\rvepsilon$-$\rvv$ & $\rvepsilon$-$\rvepsilon$ & $\rvepsilon$-$\rvx$ \\
    \midrule
    Raw joints & 1158\,$\to$\,\textbf{283.9} & 13835\,$\to$\,\textbf{272.8} & 1188\,$\to$\,\textbf{450.5} \\
    Redundant 263-dim & 359.8\,$\to$\,\textbf{74.6} & 6560\,$\to$\,\textbf{397.9} & 351.9\,$\to$\,\textbf{198.5} \\
    AE latents & 243.4\,$\to$\,\textbf{63.1} & 1185\,$\to$\,\textbf{65.3} & 168.5\,$\to$\,\textbf{93.4} \\
    \midrule
    KIT-ML ($\times10^{3}$) & \textbf{176.6}\,$\to$\,272.7 & 6029\,$\to$\,\textbf{64.3} & \textbf{166.8}\,$\to$\,527.4 \\
    \bottomrule
  \end{tabular}
\end{table}

\paragraph{Fairness Across Combinations.}
All prediction$\times$loss combinations use the same architecture, training budget, timestep sampler, solver, and number of sampling steps. They differ only where a conversion requires it at training time, and when it helps an $\rvepsilon$-prediction model. This can only improve $\rvepsilon$-prediction and thus does not impact the choice of the best model, $\rvx$-$\rvv$.

\section{Implementation Details}
\label{sec:implementation-details}

\paragraph{Datasets.}
\label{sec:datasets}
We evaluate on the HumanML3D~\cite{guo2022generating} dataset as our main benchmark. HumanML3D contains 14,616 motions from AMASS~\cite{mahmood2019amass} and HumanAct12~\cite{guo2020action2motion} with 44,970 text descriptions, at 20\,fps, each 2-10 seconds long, split into training, validation, and test sets using a 80-5-15 split. We also validate our findings on the smaller T2M benchmark of KIT-ML~\cite{plappert2016kit} containing 3,911 motions from the KIT and CMU datasets with 6,278 text descriptions, at 12.5\,fps. Both datasets are processed as in \citet{guo2022generating} with mirrored augmentations, starting at the origin facing $z+$, with the lowest joint placed on the floor. Motion sequences are capped at 196 frames.

\paragraph{Model Architecture.}
\label{sec:model-architecture}
We use the ACMDM~\cite{meng2025absolute} Transformer architecture without modification, except for additional control experiments such as adding bottleneck embeddings. Each frame of the input motion is flattened and linearly embedded as a D-dimensional token, so a motion sequence of $N$ frames becomes $N$ tokens. The same is applied to all representations, but the text follows for the raw joint positions. The backbone is a stack of DiT blocks~\cite{peebles2023scalable} with rotary positional embeddings~\cite{su2024roformer} and QK Normalizations~\cite{henry2020query} in the attention layers, and SwiGLU activations~\cite{oquab2023dinov2} in the feed-forward layers. Conditioning on the diffusion step $t$ and the text prompt is done via AdaLN~\cite{peebles2023scalable} where text is encoded by CLIP ViT-B/32~\cite{radford2021learning}, linearly projected to the hidden size, and added to the timestep embedding. In this setup, width, heads and depth are tied: a model with $L$ layers has $H = L$ heads and hidden size $D = 64\,L$. The default model we use is the S model (8 layers, 39M parameters). For additional experiments on model size, we add XS and L models (4 and 12 layers, 5M and 130M parameters). For our bottleneck experiments of \cref{tab:hml3d-bottleneck}, the linear embedding from the flattened frame to the hidden size ($J \times 3 \to D$) is replaced by a bottleneck of size $d$ which is a linear layer ($J \times 3 \to d$) followed by a linear projection to the hidden size ($d \to D$). For the latent representation, we use the pretrained autoencoder of \citet{meng2025absolute} which is a 3-block of 3-layer ResNet-based encoder and decoder architecture, downsampling the motion sequence by a factor of 4 in time.

\paragraph{Training Details.}
\label{sec:training-details}
For training, we follow ACMDM choices, including AdamW optimizer with $\beta_1 = 0.9$, $\beta_2 = 0.99$, weight decay of $10^{-5}$, a learning rate of $2 \times 10^{-4}$, and a $10\times$ decay at 50K iterations, a batch size of 64, and we train for 500 epochs (191.5K iterations) on HumanML3D. On KIT-ML, we train for 55K iterations to avoid overfitting. An exponential moving average (EMA) of the model weights is maintained with a decay of 0.9999, which is later used for evaluation. The prediction output, loss space, and $t$-sampler are the settings that change between runs. The loss weights of \cref{tab:loss-weights} are applied with $t$ and $1-t$ floored at $\delta = 0.05$ to stabilize training. The logit-normal sampler uses $\mu = -0.8$ and $\sigma = 0.8$ unless stated otherwise. Training our default S model on the HumanML3D dataset with a single NVIDIA H100 requires 8.8 hours with the above settings. The exact configurations used to train our top recipe is summarized in \cref{tab:training-config}

\begin{table}[t]
  \caption{Training configurations of our recipe.}
  \label{tab:training-config}
  \centering
  \footnotesize
  \setlength{\tabcolsep}{4pt}
  \begin{tabular}{@{}lc@{}}
    \toprule
    & \\
    \midrule
    Optimizer          & AdamW ($\beta_1{=}0.9$, $\beta_2{=}0.99$, weight decay $10^{-5}$) \\
    Learning rate      & $2\times10^{-4}$ \\
    Warm-up            & linear, 2K iterations \\
    LR schedule        & step decay ($\times 0.1$) at 50K iterations \\
    Batch size         & 64 \\
    Training length    & 500 epochs (191.5K iterations) \\
    EMA decay          & 0.9999 \\
    Denominator clip $\delta$ & 0.05 \\
    CFG $p_\text{uncond}$ & 0.1 \\
    \bottomrule
  \end{tabular}
\end{table}

\paragraph{Evaluation Metrics.}
\label{sec:evaluation-metrics}
We run our evaluations on the EMA weights, using a 50-step Euler sampler, and CFG scale $s=3.0$ unless stated otherwise, in batches of 32 captions from the test set with one motion generated per caption at the ground-truth length. Every reported number is the mean over 20 evaluation repetitions with its 95\% confidence interval unless stated otherwise. Generated motions of any representation are first mapped to absolute joint positions before metric computation. 

\emph{\textbf{T2M metrics.}} We compute the T2M metrics of \citet{guo2022generating} using a text feature extractor and a motion feature extractor jointly trained under a contrastive loss to produce geometrically close feature vectors for matched text-motion pairs, and vise versa. On HumanML3D, we use the extractors retrained by ACMDM~\cite{meng2025absolute} on the 67-dimensional joint features of \citet{meng2025rethinking}, on KIT-ML we use the original evaluators. See \cref{supp:evaluator} for discussion on choice of evaluator. \emph{FID}~\cite{heusel2017gans} (Fr\'echet Inception Distance) measures how closely the distribution of generated motions matches the distribution of real test-set motions. Using a motion-feature extractor, we fit a Gaussian distribution to the feature embeddings of real and generated test motions and compute the Fr\'echet distance between the two distributions. \emph{R-Precision} measures how well a generated motion matches its text prompt. For each generated motion in a 32-sample evaluation batch, we rank all 32 captions in the batch by their feature-space distance to the motion, and report how often the ground-truth caption ranks in the top 1, 2, or 3. \emph{MM-Dist} (Multimodal Distance) measured the alignment between generated motions and their text prompts. It computes the mean Euclidean distance in the evaluator's joint embedding space between each generated motion and its corresponding input caption, where lower values indicate tighter text--motion correspondence. \emph{Diversity} measures how different the generated motions are across the entire test distribution. It computes the mean Euclidean distance between the feature embeddings of 300 random pairs of generated motions, with optimal values matching the diversity score of real test motions. \emph{MultiModality} measures how many distinct motions a model can produce for the exact same text prompt. It generates 30 motions for each of 96 test captions and averages the feature-space Euclidean distance over 10 random sampled pairs per caption.

\begin{algorithm}[t]
  \caption{Training}
  \label{alg:training}
  \begin{algorithmic}[1]
    \Require output $\rvu$, loss space $\boldsymbol{\ell}$, timestep sampler $p(t)$
    \Repeat
      \State $(\rvx, \rvc) \sim q(\rvx, \rvc)$
        \Comment{motion and text}
      \State $\rvc \gets \varnothing$ with probability $p_\text{uncond}$
      \State $t \sim p(t)$, $\rvepsilon \sim \mathcal{N}(\mathbf{0}, \mathbf{I})$
      \State $\rvz_t \gets t\,\rvx + (1-t)\,\rvepsilon$
      \State $\rvu \gets \rvx - \rvepsilon$, $\rvepsilon$, or $\rvx$
        \Comment{\cref{tab:parameterization-targets}}
      \State Take gradient descent step on
      \Statex \hspace{\algorithmicindent}
        $\nabla_\theta\, w_{\rvu \to \boldsymbol{\ell}}(t)\,
         \big\| \rvu_\theta(\rvz_t, t, \rvc) - \rvu \big\|^2$
        \Comment{\cref{tab:loss-weights}; \cref{supp:flooring}}
    \Until{converged}
  \end{algorithmic}
\end{algorithm}

\begin{algorithm}[t]
  \caption{Sampling}
  \label{alg:sampling}
  \begin{algorithmic}[1]
    \Require output $\rvu$, text $\rvc$, guidance scale $s$, steps $N$, interval $[t_0, t_1]$
    \State $\rvz_0 \sim \mathcal{N}(\mathbf{0}, \mathbf{I})$
    \State $h = (t_1 - t_0) / N$
    \For{$t = t_0, t_0+h, \dots, t_1-h$}
      \State $\rvu_\theta \gets \rvu_\theta(\rvz_t, t, \varnothing)
              + s\,\big(\rvu_\theta(\rvz_t, t, \rvc) - \rvu_\theta(\rvz_t, t, \varnothing)\big)$
      \State $\hat{\rvv} \gets$ velocity recovered from $\rvu_\theta$
        \Comment{\cref{tab:conversions}, \cref{supp:flooring}}
      \State $\rvz_{t+h} \gets \rvz_t + h\,\hat{\rvv}$
    \EndFor
    \State \Return $\rvz_{t_1}$
      \Comment{generated motion}
  \end{algorithmic}
\end{algorithm}

\emph{\textbf{Plausibility metrics.}} We observe that T2M metrics are nearly blind to visual quality of motions. A model with very low FID can produce samples that demonstrate significant jitter (see supplementary video). The feature extractors used to compute the T2M metrics are trained to match motion to a caption, and thus these artefacts have minimal impact on the embeddings. We therefore add four metrics as proxy measures of jitter in the generated joint positions $\mathbf{p}_n$, averaged over joints, frames and motions, and compare each with its value on the real test set motions. \emph{Acceleration}~\cite{rempe2021humor, yuan2022glamr, ye2023decoupling} is the mean norm of the second finite difference $\|\mathbf{p}_{n+1} - 2\mathbf{p}_n + \mathbf{p}_{n-1}\|\,f^2$, with $f$ being the frame rate. \emph{Jerk}~\cite{ye2023decoupling, shin2024wham} is the mean norm of the third finite difference $\|\mathbf{p}_{n+2} - 3\mathbf{p}_{n+1} + 3\mathbf{p}_n - \mathbf{p}_{n-1}\|\,f^3$. Both should consider the ground truth data as well, so values closer to those of the real motions are better. \emph{Bone-length Change} is the mean absolute frame-to-frame change in bone length $\big|\,\|\mathbf{p}^a_{n+1} - \mathbf{p}^b_{n+1}\| - \|\mathbf{p}^a_n - \mathbf{p}^b_n\|\,\big|$, where a bone $(a, b)$ is a pair of adjacent joints along the kinematic chains of the skeleton (21 bones on HumanML3D, 20 on KIT-ML). It is zero for real motions, whose bone lengths are fixed. \emph{Floor Penetration} is the mean of $\max(0,-y)$ over all joints and frames, with the floor at $y=0$; it is zero for real motions. HumanML3D metrics are computed in m/s$^2$ (Acceleration) and m/s$^3$ (Jerk) at 20\,fps, and mm/frame (Bone-length Change) and mm (Floor Penetration). KIT-ML keeps its native units $u$ at 12.5\,fps.

\section{Discussion}

\subsection{Choice of Evaluator}
\label{supp:evaluator}
Most diffusion-based human motion generation models (\eg MDM and MLD) use the standard HumanML3D evaluator of \citet{guo2022generating} which takes the 263-dimensional relative-local redundant representation as input. Our models instead generate the $22 \times 3$-dimensional joint positions, without rotations, velocities or foot contacts. To use the standard evaluator, it was required to run inverse kinematics to compute joint rotations, finite differences for velocities, and apply a threshold for foot contacts. As a result, the evaluation score would measure this conversion as well as the model. Therefore, we follow \citet{meng2025rethinking, meng2025absolute} and use their evaluator, which has the same architecture and objective but takes only the 67 dimensions determined by the joint positions (root motion and joint positions in local root coordinates). This means that our reported metrics computed on HumanML3D are comparable with numbers reported by \citet{meng2025rethinking, meng2025absolute}, but not with numbers reported under the 263-dimensional evaluator. On KIT-ML we use the standard evaluator, with its input features computed from the generated joints.

\subsection{Logit-Normal $t$-Sampler}
\label{sec:supp-tsampler}
\begin{figure}[t]
  \centering
  \includegraphics[width=0.86\linewidth]{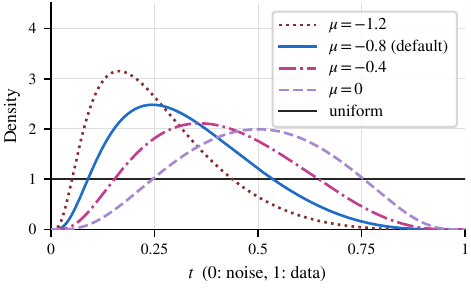}
  \caption{Density of the logit-normal $t$-sampler, $t = \sigma(\xi)$ with $\xi \sim \mathcal{N}(\mu, 0.8^2)$, for $\mu \in \{-1.2, -0.8, -0.4, 0\}$, compared with the uniform sampler.}
  \label{fig:t-sampler}
\end{figure}

Our recommended recipe replaces the uniform training distribution over $t$ with the logit-normal sampler $t = \mathrm{sigmoid}(\xi)$ with $\xi \sim \mathcal{N}(\mu, \sigma^2)$, using $\mu=-0.8$ and $\sigma=0.8$ that \citet{li2026back} use for raw image pixels. Compared to a uniform $t$-sampler, this concentrates training on noisy inputs (\cref{fig:t-sampler}).

\paragraph{Effect on combinations.}
The choice of sampler has opposite effects on $\rvv$-$\rvv$ and $\rvx$-$\rvv$ (\cref{tab:hml3d-tsampler}). With $\rvx$-prediction, both FID and Jerk improve while with $\rvv$-prediction, the same sampler is catastrophic and raises FID from 0.149 to 13.995 and Jerk from 120.2 to 660.0\,m/s$^3$. Since both models compute the loss on the velocity, the difference stems from less training as $t \to 1$. With $\rvx$-prediction, the target is close to the network input $\rvz_t \approx \rvx$ and little training is required. With $\rvv$-prediction, the prediction target $\rvv = \rvx - \rvepsilon$ requires recovering the noise from the almost clean-sample input, which is a much harder task.

\paragraph{Choice of $\mu$.}
The value of $\mu=-0.8$ was chosen for images, so we sweep $\mu \in \{-1.2, -0.8, -0.4, 0\}$ for the $\rvx - \rvv$ model (\cref{tab:hml3d-mu}). The gain in smoothness does not depend on $\mu$: jerk stays between 62.4 and 64.1\,m/s$^3$ for all four values, against 83.0 with uniform $t$. FID is lowest at $\mu=-0.8$ and R-Precision is highest at $\mu=-0.8$, so we keep the default without further tuning.

\paragraph{Limitations.}
The logit-normal sampler lowers text-alignment, as seen in the drop of R-Precision values from Uniform to every $\mu$ in \cref{tab:hml3d-mu}. We keep the sampler that gains in FID, Diversity, Acceleration, and Jerk, as text-alignment is not our top-priority but perceptual motion quality and plausibility is. We expect that a sweep over CFG scale can help adjust text alignment at inference time.

\begin{table}[t]
  \caption{Effect of the training $t$-sampler on HumanML3D, for $\rvv$-prediction and $\rvx$-prediction, both with the $\rvv$-loss.
  The logit-normal sampler ($\mu=-0.8$, $\sigma=0.8$) concentrates training on intermediate noise levels, whereas uniform $t$ covers all noise levels equally.
  All other settings match \cref{tab:hml3d-3x3} of the paper; we report the mean and, as a superscript, the 95\% confidence interval over 20 evaluation repeats.
  $\rightarrow$: closer to Real is better.
  Bold marks the best model in each column.}
  \label{tab:hml3d-tsampler}
  \centering
  \footnotesize
  \setlength{\tabcolsep}{2pt}
  \begin{tabular}{@{}llcccc@{}}
    \toprule
    & & \multicolumn{3}{c}{Text-to-Motion metrics} & \shortstack{Physical\\plausibility} \\
    \cmidrule(lr){3-5} \cmidrule(l){6-6}
    Model & $t$-sampler & FID$\downarrow$ & Top 1$\uparrow$ & MModality$\uparrow$ & Jerk$\rightarrow$ \\
    & & & & & m/s$^3$ \\
    \midrule
    \multicolumn{2}{@{}l}{Real} & -- & 0.500$^{\pm.002}$ & -- & 48.9$^{\pm0.0}$ \\
    \midrule
    $\rvv$-$\rvv$ & uniform & 0.149$^{\pm.007}$ & 0.499$^{\pm.003}$ & 2.165$^{\pm.076}$ & 120.2$^{\pm0.4}$ \\
    $\rvv$-$\rvv$ & logit-normal & 13.995$^{\pm.066}$ & 0.308$^{\pm.002}$ & 2.109$^{\pm.077}$ & 660.0$^{\pm0.8}$ \\
    \midrule
    $\rvx$-$\rvv$ & uniform & 0.134$^{\pm.004}$ & \textbf{0.508}$^{\pm.003}$ & 2.057$^{\pm.081}$ & 83.0$^{\pm0.2}$ \\
    $\rvx$-$\rvv$ & logit-normal & \textbf{0.087}$^{\pm.006}$ & 0.475$^{\pm.003}$ & \textbf{2.239}$^{\pm.078}$ & \textbf{62.4}$^{\pm0.1}$ \\
    \bottomrule
  \end{tabular}
\end{table}

\begin{table*}[t]
  \caption{Mean $\mu$ of the logit-normal $t$-sampler on HumanML3D, for $\rvx$-prediction with the $\rvv$-loss and $\sigma=0.8$. Real shows the real test set metrics and Uniform shows the values for the $\rvx$-$\rvv$ model trained with the uniform $t$-sampler.
  All other settings match \cref{tab:hml3d-tsampler}; we report the mean and, as a superscript, the 95\% confidence interval over 20 evaluation repeats.
  $\downarrow$: lower is better, $\uparrow$: higher is better, $\rightarrow$: closer to Real is better.
  Bold marks the best model in each column.}
  \label{tab:hml3d-mu}
  \centering
  \footnotesize
  \setlength{\tabcolsep}{2.8pt}
  \begin{tabular}{@{}ccccccccccccc@{}}
    \toprule
    & \multicolumn{7}{c}{Text-to-Motion metrics} & \multicolumn{4}{c}{Physical plausibility} \\
    \cmidrule(lr){2-8} \cmidrule(l){9-12}
    & & \multicolumn{3}{c}{R-Precision$\uparrow$} & & & & & & & \\
    \cmidrule(lr){3-5}
    $\mu$ & FID$\downarrow$ & Top 1 & Top 2 & Top 3 & MM-Dist$\downarrow$ & Diversity$\rightarrow$ & MModality$\uparrow$
      & Bone$\downarrow$ & Accel.$\rightarrow$ & Jerk$\rightarrow$ & Floor$\downarrow$ \\
    & & & & & & & & mm/frame & m/s$^2$ & m/s$^3$ & mm \\
    \midrule
    Real
      & -- & 0.500$^{\pm.002}$ & 0.694$^{\pm.003}$ & 0.793$^{\pm.002}$ & 3.293$^{\pm.008}$ & 10.506$^{\pm.105}$ & --
      & 0.000$^{\pm.000}$ & 2.764$^{\pm.001}$ & 48.9$^{\pm0.0}$ & 0.000$^{\pm.000}$ \\
    \midrule
    Uniform
      & {0.134}$^{\pm.004}$ & \textbf{0.508}$^{\pm.003}$ & \textbf{0.700}$^{\pm.003}$ & \textbf{0.797}$^{\pm.003}$ & \textbf{3.251}$^{\pm.010}$ & 10.803$^{\pm.075}$ & 2.057$^{\pm.081}$
      & 1.573$^{\pm.003}$ & 3.646$^{\pm.010}$ & 83.0$^{\pm0.2}$ & 0.055$^{\pm.003}$ \\
    \midrule
    $-1.2$
      & 0.137$^{\pm.007}$ & 0.466$^{\pm.002}$ & 0.658$^{\pm.003}$ & 0.760$^{\pm.002}$ & 3.471$^{\pm.009}$ & 10.235$^{\pm.080}$ & 2.056$^{\pm.075}$
      & 2.140$^{\pm.005}$ & 3.089$^{\pm.010}$ & 62.5$^{\pm0.1}$ & 0.122$^{\pm.002}$ \\
    $-0.8$
      & \textbf{0.087}$^{\pm.006}$ & {0.475}$^{\pm.003}$ & {0.665}$^{\pm.003}$ & {0.766}$^{\pm.002}$ & {3.447}$^{\pm.012}$ & \textbf{10.507}$^{\pm.073}$ & 2.239$^{\pm.078}$
      & 1.964$^{\pm.003}$ & \textbf{3.070}$^{\pm.006}$ & \textbf{62.4}$^{\pm0.1}$ & 0.065$^{\pm.001}$ \\
    $-0.4$
      & 0.113$^{\pm.008}$ & 0.462$^{\pm.003}$ & 0.652$^{\pm.002}$ & 0.753$^{\pm.002}$ & 3.552$^{\pm.012}$ & 10.690$^{\pm.082}$ & 2.658$^{\pm.094}$
      & 1.841$^{\pm.004}$ & 3.153$^{\pm.008}$ & 64.1$^{\pm0.1}$ & 0.041$^{\pm.001}$ \\
    $0.0$
      & 0.123$^{\pm.008}$ & 0.427$^{\pm.003}$ & 0.612$^{\pm.003}$ & 0.715$^{\pm.003}$ & 3.837$^{\pm.012}$ & 10.723$^{\pm.082}$ & \textbf{3.218}$^{\pm.129}$
      & \textbf{1.674}$^{\pm.003}$ & 3.083$^{\pm.007}$ & 63.4$^{\pm0.1}$ & \textbf{0.040}$^{\pm.001}$ \\
    \bottomrule
  \end{tabular}
\end{table*}

\begin{table*}[t]
  \caption{Comparison with prior text-to-motion diffusion methods on HumanML3D. Repr.: the representation each model operates on (Redund.: the 263-dimensional HumanML3D~\cite{guo2022generating} features; Latent: autoencoder latents; Raw: joint positions). Pred.: the network output ($\rvx$: clean motion, $\rvepsilon$: noise, $\rvv$: velocity).
  All models are evaluated with the evaluator of \citet{meng2025rethinking, meng2025absolute} (\cref{supp:evaluator}), and the physical-plausibility metrics are computed on the joint positions recovered from each representation.
  We report the mean and, as a superscript, the 95\% confidence interval over 20 evaluation repeats.
  $\downarrow$: lower is better, $\uparrow$: higher is better, $\rightarrow$: closer to Real is better.
  Bold marks the best model in each column, except for multimodality.}
  \label{tab:prior-methods}
  \centering
  \footnotesize
  \setlength{\tabcolsep}{1.3pt}
  \resizebox{\textwidth}{!}{%
  \begin{tabular}{@{}lllccccccccccc@{}}
    \toprule
    & & & \multicolumn{7}{c}{Text-to-Motion metrics} & \multicolumn{4}{c}{Physical plausibility} \\
    \cmidrule(lr){4-10} \cmidrule(l){11-14}
    & & & & \multicolumn{3}{c}{R-Precision$\uparrow$} & & & & & & & \\
    \cmidrule(lr){5-7}
    Model & Repr. & Pred. & FID$\downarrow$ & Top 1 & Top 2 & Top 3 & MM-Dist$\downarrow$ & Diversity$\rightarrow$ & MModality$\uparrow$
      & Bone$\downarrow$ & Accel.$\rightarrow$ & Jerk$\rightarrow$ & Floor$\downarrow$ \\
    & & & & & & & & & & mm/frame & m/s$^2$ & m/s$^3$ & mm \\
    \midrule
    Real & -- & --
      & -- & 0.500$^{\pm.002}$ & 0.694$^{\pm.003}$ & 0.793$^{\pm.002}$ & 3.293$^{\pm.008}$ & 10.506$^{\pm.105}$ & --
      & 0.000$^{\pm.000}$ & 2.764$^{\pm.001}$ & 48.9$^{\pm0.0}$ & 0.000$^{\pm.000}$ \\
    \midrule
    MDM~\cite{tevet2022human} & Redund. & $\rvx$
      & 0.410$^{\pm.011}$ & 0.438$^{\pm.002}$ & 0.631$^{\pm.003}$ & 0.738$^{\pm.002}$ & 3.674$^{\pm.010}$ & 10.897$^{\pm.112}$ & 2.352$^{\pm.086}$
      & 0.612$^{\pm.002}$ & 2.322$^{\pm.008}$ & 39.2$^{\pm0.2}$ & \textbf{0.016}$^{\pm.000}$ \\
    MDM (raw joints) & Raw & $\rvx$
      & 1.557$^{\pm.028}$ & 0.371$^{\pm.003}$ & 0.552$^{\pm.003}$ & 0.660$^{\pm.004}$ & 4.229$^{\pm.016}$ & 10.773$^{\pm.112}$ & 3.138$^{\pm.112}$
      & 0.965$^{\pm.003}$ & 2.273$^{\pm.013}$ & 42.5$^{\pm0.3}$ & 0.039$^{\pm.001}$ \\
    MotionDiffuse~\cite{zhang2024motiondiffuse} & Redund. & $\rvepsilon$
      & 0.945$^{\pm.015}$ & 0.461$^{\pm.003}$ & 0.650$^{\pm.003}$ & 0.754$^{\pm.002}$ & 3.487$^{\pm.011}$ & 10.432$^{\pm.074}$ & 1.756$^{\pm.076}$
      & \textbf{0.417}$^{\pm.001}$ & 1.974$^{\pm.005}$ & 40.1$^{\pm0.1}$ & 0.029$^{\pm.001}$ \\
    MLD++~\cite{chen2023executing, dai2024motionlcm} & Latent & $\rvepsilon$
      & 2.310$^{\pm.028}$ & 0.494$^{\pm.003}$ & 0.687$^{\pm.002}$ & 0.784$^{\pm.002}$ & 3.265$^{\pm.009}$ & 9.379$^{\pm.071}$ & 1.857$^{\pm.061}$
      & 0.735$^{\pm.003}$ & \textbf{2.777}$^{\pm.008}$ & \textbf{53.9}$^{\pm0.2}$ & 0.028$^{\pm.002}$ \\
    MotionLCM~\cite{dai2024motionlcm} & Latent & $\rvepsilon$
      & 2.579$^{\pm.033}$ & 0.496$^{\pm.003}$ & 0.690$^{\pm.003}$ & 0.788$^{\pm.003}$ & \textbf{3.222}$^{\pm.010}$ & 9.345$^{\pm.085}$ & 2.032$^{\pm.063}$
      & 0.895$^{\pm.004}$ & 3.453$^{\pm.011}$ & 74.7$^{\pm0.3}$ & 0.049$^{\pm.002}$ \\
    ACMDM~\cite{meng2025absolute} & Latent & $\rvv$
      & 0.109$^{\pm.004}$ & \textbf{0.508}$^{\pm.002}$ & \textbf{0.701}$^{\pm.003}$ & \textbf{0.798}$^{\pm.003}$ & 3.254$^{\pm.010}$ & 10.807$^{\pm.091}$ & 2.150$^{\pm.097}$
      & 2.110$^{\pm.007}$ & 3.140$^{\pm.009}$ & 67.4$^{\pm0.2}$ & 0.092$^{\pm.003}$ \\
    ACMDM (raw joints) & Raw & $\rvv$
      & 0.134$^{\pm.006}$ & 0.497$^{\pm.003}$ & 0.691$^{\pm.002}$ & 0.789$^{\pm.003}$ & 3.312$^{\pm.010}$ & 10.720$^{\pm.080}$ & 2.243$^{\pm.077}$
      & 2.095$^{\pm.011}$ & 4.160$^{\pm.013}$ & 103.4$^{\pm0.5}$ & 0.078$^{\pm.004}$ \\
    Ours & Raw & $\rvx$
      & \textbf{0.087}$^{\pm.006}$ & 0.475$^{\pm.003}$ & 0.665$^{\pm.003}$ & 0.766$^{\pm.002}$ & 3.447$^{\pm.012}$ & \textbf{10.507}$^{\pm.073}$ & 2.239$^{\pm.078}$
      & 1.964$^{\pm.003}$ & 3.070$^{\pm.006}$ & 62.4$^{\pm0.1}$ & 0.065$^{\pm.001}$ \\
    \bottomrule
  \end{tabular}}
\end{table*}

\section{Additional Results}
\label{sec:additional-results}
This section provides the complete version of results reported in compact form in the paper or earlier in this appendix, and compares our recipe with prior T2M methods. Each table reports all T2M and physical-plausibility metrics, as the mean and 95\% confidence interval over 20 evaluation repeats, under the protocol of \cref{tab:hml3d-3x3} of the paper unless stated otherwise.

\subsection{Comparison with Prior Methods}
\label{supp:comparison-with-prior}
Our goal in this work is not to propose a new T2M model that is SOTA on every metric, but rather to obtain a clear understanding of what matters in motion diffusion when operating on raw joint position data, in terms of the prediction output, the loss space, and closely related choices such as the $t$-sampler. For completeness, \cref{tab:prior-methods} compares our recipe against commonly used T2M methods. We evaluate every method with the evaluator of \citet{meng2025rethinking} (\cref{supp:evaluator}), since the numbers reported in the original papers use a different evaluator and are therefore not directly transferable. For Acceleration and Jerk, the reference values come from the real test set. Methods below the real value are over-smoothed, methods above it are jittery, and both are considered as deviations from real motion.

\paragraph{Key takeaways.}
Our recipe has the lowest FID and matches the Diversity of the real test set. Among the three models with FID below $0.2$ (ours and the two ACMDM variants), it is also the closest to real motion in Acceleration and Jerk, without an autoencoder. MDM~\cite{tevet2022human} and MotionDiffuse~\cite{zhang2024motiondiffuse} are over-smoothed and have considerably higher FID. MLD++~\cite{dai2024motionlcm} is the closest to real motion in Acceleration and Jerk overall, but has much higher FID and a lower Diversity.

\paragraph{MDM on raw joints.}
The original MDM model operates on the redundant HumanML3D representation and is an $x$-prediction, $x$-loss model under the DDPM paradigm. We additionally train it on raw joint positions here. Its FID becomes much worse, and its Jerk rises only slightly (still below real value).

\subsection{Top Recipes}
\Cref{tab:hml3d-recipes} is the full version of \cref{tab:hml3d-recipes-compact} of the paper.
It compares the $\rvv$-$\rvv$ and $\rvx$-$\rvv$ models sampled with 50 Euler steps, the ACMDM recipe (the same $\rvv$-$\rvv$ model sampled with Dopri5), and our recipe ($\rvx$-$\rvv$ with the logit-normal $t$-sampler, 50 Euler steps) on every metric.

\begin{table*}[t]
  \caption{$\rvx$-$\rvv$ and $\rvv$-$\rvv$ combinations of \cref{tab:hml3d-3x3} of the paper vs. the recipe of ACMDM~\cite{meng2025absolute} (Baseline) and our recommended recipe (Ours). Top: the $\rvv$-$\rvv$ and $\rvx$-$\rvv$ are trained with uniform $t$ and sampled with 50 Euler steps.
  Bottom: the recipe of ACMDM, which samples the same $\rvv$-$\rvv$ model with the adaptive Dopri5 solver, and our recipe which trains the $\rvx$-$\rvv$ model with the logit-normal $t$-sampler ($\mu=-0.8$) and samples it with 50 Euler steps. We report the mean and, as a superscript, the 95\% confidence interval over 20 evaluation repeats.
  $\downarrow$: lower is better, $\uparrow$: higher is better, $\rightarrow$: closer to the real test motions is better (Diversity 10.506, Acceleration 2.764\,m/s$^2$, Jerk 48.9\,m/s$^3$).
  Bold marks the best model in each column, except for multimodality.}
  \label{tab:hml3d-recipes}
  \centering
  \footnotesize
  \setlength{\tabcolsep}{1.3pt}
  \begin{tabular}{@{}lccccccccccc@{}}
    \toprule
    & \multicolumn{7}{c}{Text-to-Motion metrics} & \multicolumn{4}{c}{Physical plausibility} \\
    \cmidrule(lr){2-8} \cmidrule(l){9-12}
    & & \multicolumn{3}{c}{R-Precision$\uparrow$} & & & & & & & \\
    \cmidrule(lr){3-5}
    Model & FID$\downarrow$ & Top 1 & Top 2 & Top 3 & MM-Dist$\downarrow$ & Diversity$\rightarrow$ & MModality$\uparrow$
      & Bone$\downarrow$ & Accel.$\rightarrow$ & Jerk$\rightarrow$ & Floor$\downarrow$ \\
    & & & & & & & & mm/frame & m/s$^2$ & m/s$^3$ & mm \\
    \midrule
    $\rvv$-$\rvv$
      & 0.149$^{\pm.007}$ & 0.499$^{\pm.003}$ & 0.693$^{\pm.003}$ & 0.791$^{\pm.003}$ & 3.295$^{\pm.010}$ & 10.725$^{\pm.078}$ & 2.165$^{\pm.076}$
      & 2.712$^{\pm.009}$ & 4.457$^{\pm.011}$ & 120.2$^{\pm0.4}$ & 0.076$^{\pm.004}$ \\
    $\rvx$-$\rvv$
      & 0.134$^{\pm.004}$ & \textbf{0.508}$^{\pm.003}$ & \textbf{0.700}$^{\pm.003}$ & \textbf{0.797}$^{\pm.003}$ & \textbf{3.251}$^{\pm.010}$ & 10.803$^{\pm.075}$ & 2.057$^{\pm.081}$
      & \textbf{1.573}$^{\pm.003}$ & 3.646$^{\pm.010}$ & 83.0$^{\pm0.2}$ & \textbf{0.055}$^{\pm.003}$ \\
    \midrule
    Baseline
      & 0.134$^{\pm.006}$ & 0.497$^{\pm.003}$ & 0.691$^{\pm.002}$ & 0.789$^{\pm.003}$ & 3.312$^{\pm.010}$ & 10.720$^{\pm.080}$ & 2.243$^{\pm.077}$
      & 2.095$^{\pm.011}$ & 4.160$^{\pm.013}$ & 103.4$^{\pm0.5}$ & 0.078$^{\pm.004}$ \\
    Ours
      & \textbf{0.087}$^{\pm.006}$ & 0.475$^{\pm.003}$ & 0.665$^{\pm.003}$ & 0.766$^{\pm.002}$ & 3.447$^{\pm.012}$ & \textbf{10.507}$^{\pm.073}$ & 2.239$^{\pm.078}$
      & 1.964$^{\pm.003}$ & \textbf{3.070}$^{\pm.006}$ & \textbf{62.4}$^{\pm0.1}$ & 0.065$^{\pm.001}$ \\
    \bottomrule
  \end{tabular}
\end{table*}

\subsection{KIT-ML}
\begin{figure*}[t]
  \centering
  \includegraphics[width=0.85\linewidth]{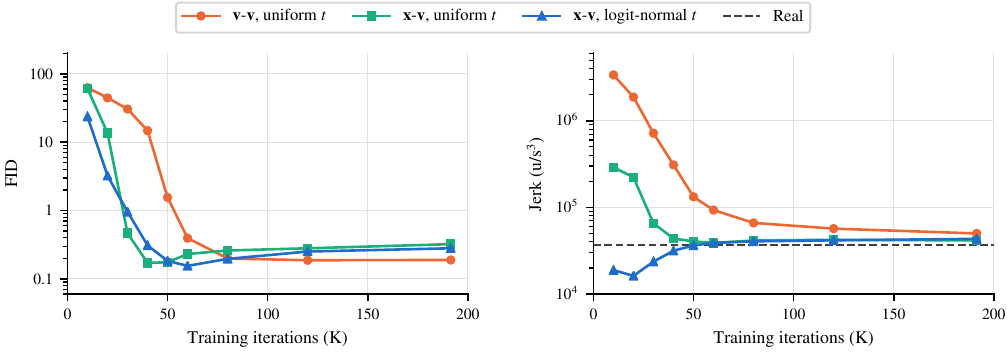}
  \begin{subfigure}{0.49\linewidth}
    \centering
    \caption{FID$\downarrow$}
    \label{fig:kit-curves-fid}
  \end{subfigure}
  \hfill
  \begin{subfigure}{0.49\linewidth}
    \centering
    \caption{Jerk$\rightarrow$ in the native units of KIT-ML; the dashed line is the jerk of the real test motions.}
    \label{fig:kit-curves-jerk}
  \end{subfigure}
  \caption{Training curves of the S models on KIT-ML, trained for 191.5K iterations (about 2800 epochs) with the learning rate decayed by $10\times$ at 50K iterations.
  Each point is a checkpoint sampled with 50 Euler steps and evaluated with 5 repeats.
  The $\rvx$-prediction models reach their best FID at 40--60K iterations and degrade afterwards, whereas $\rvv$-$\rvv$ keeps improving until about 120K; \cref{tab:kit-3x3} therefore reports a 55K-iteration budget.}
  \label{fig:kit-curves}
\end{figure*}

\Cref{tab:kit-3x3} reports all metrics for the nine prediction$\times$loss combinations models on KIT-ML dataset. \Cref{tab:kit-grid-combined} of the paper shows only the FID and Jerk of these results as a summary. Similar to HumanML3D results, $\rvx$-$\rvv$ also outperforms on R-Precision and MM-Dist. \Cref{fig:kit-curves} shows the training curves of KIT-ML models, trained beyond their set budget of 55K-iterations. $\rvx$-prediction models reach their best FID at 40--60K iterations and degrade afterwards, while $\rvv$-$\rvv$ keeps improving until about 120K iterations.

\begin{table*}[t]
  \caption{Prediction target \vs loss space on KIT-ML.
  Same protocol as \cref{tab:hml3d-3x3} of the paper (flow matching on raw joint positions, uniform $t$-sampler, 50 Euler steps, classifier-free guidance scale 3.0), but trained for 55K iterations with the learning rate decayed at 45K; KIT-ML has $4.8\times$ fewer training clips than HumanML3D, so this budget corresponds to about 800 epochs.
  Metrics use the standard KIT-ML evaluator, and the physical-plausibility metrics are in the native units of KIT-ML (u, millimetres of the 1\,m tall reference skeleton), so they are not comparable to the HumanML3D values.
  We report the mean and, as a superscript, the 95\% confidence interval over 20 evaluation repeats.
  $\downarrow$: lower is better, $\uparrow$: higher is better, $\rightarrow$: closer to Real is better.
  For $\rvepsilon$-prediction we report the better in FID of exact and floored sampling; $\dagger$ marks floored sampling (\cref{tab:floor-fid}).
  Bold marks the best model in each column.}
  \label{tab:kit-3x3}
  \centering
  \footnotesize
  \setlength{\tabcolsep}{2.2pt}
  \begin{tabular}{@{}cccccccccccc@{}}
    \toprule
    & & \multicolumn{6}{c}{Text-to-Motion metrics} & \multicolumn{4}{c}{Physical plausibility} \\
    \cmidrule(lr){3-8} \cmidrule(l){9-12}
    & & & \multicolumn{3}{c}{R-Precision$\uparrow$} & & & & & & \\
    \cmidrule(lr){4-6}
    Pred. & Loss & FID$\downarrow$ & Top 1 & Top 2 & Top 3 & MM-Dist$\downarrow$ & Diversity$\rightarrow$
      & Bone$\downarrow$ & Accel.$\rightarrow$ & Jerk$\rightarrow$ & Floor$\downarrow$ \\
    & & & & & & & & u/frame & u/s$^2$ & u/s$^3$ & u \\
    \midrule
    \multicolumn{2}{c}{Real}
      & -- & 0.424$^{\pm.006}$ & 0.648$^{\pm.008}$ & 0.777$^{\pm.006}$ & 2.810$^{\pm.016}$ & 10.999$^{\pm.104}$
      & 0.000$^{\pm.000}$ & 2480$^{\pm6}$ & 37302$^{\pm95}$ & 0.000$^{\pm.000}$ \\
    \midrule
    $\rvv$ & $\rvv$
      & 1.139$^{\pm.035}$ & 0.391$^{\pm.006}$ & 0.605$^{\pm.006}$ & 0.738$^{\pm.007}$ & 3.251$^{\pm.027}$ & 10.103$^{\pm.088}$
      & 10.22$^{\pm.06}$ & 5853$^{\pm26}$ & 122296$^{\pm620}$ & 0.005$^{\pm.001}$ \\
    $\rvv$ & $\rvepsilon$
      & 0.511$^{\pm.023}$ & 0.384$^{\pm.006}$ & 0.596$^{\pm.005}$ & 0.723$^{\pm.006}$ & 3.204$^{\pm.036}$ & 10.516$^{\pm.099}$
      & 8.021$^{\pm.034}$ & 4638$^{\pm16}$ & 93832$^{\pm351}$ & 0.014$^{\pm.001}$ \\
    $\rvv$ & $\rvx$
      & 26.32$^{\pm.21}$ & 0.214$^{\pm.005}$ & 0.363$^{\pm.004}$ & 0.478$^{\pm.005}$ & 6.468$^{\pm.027}$ & 5.890$^{\pm.053}$
      & 40.25$^{\pm.08}$ & 19542$^{\pm39}$ & 442316$^{\pm893}$ & 0.018$^{\pm.001}$ \\
    \midrule
    $\rvepsilon$ & $\rvv$
      & 0.782$^{\pm.055}$ & 0.408$^{\pm.007}$ & 0.620$^{\pm.006}$ & 0.741$^{\pm.004}$ & 3.238$^{\pm.019}$ & 10.761$^{\pm.100}$
      & 16.49$^{\pm1.95}$ & 8548$^{\pm883}$ & 176562$^{\pm20337}$ & 0.677$^{\pm.119}$ \\
    $\rvepsilon$\floored & $\rvepsilon$
      & 0.505$^{\pm.029}$ & 0.381$^{\pm.007}$ & 0.596$^{\pm.007}$ & 0.726$^{\pm.005}$ & 3.342$^{\pm.025}$ & 10.837$^{\pm.097}$
      & 5.208$^{\pm.028}$ & 3559$^{\pm13}$ & 64300$^{\pm258}$ & 0.137$^{\pm.005}$ \\
    $\rvepsilon$ & $\rvx$
      & 2.440$^{\pm.108}$ & 0.355$^{\pm.007}$ & 0.553$^{\pm.006}$ & 0.672$^{\pm.006}$ & 3.927$^{\pm.030}$ & 10.125$^{\pm.091}$
      & 16.58$^{\pm1.71}$ & 7822$^{\pm745}$ & 166844$^{\pm17019}$ & 0.432$^{\pm.208}$ \\
    \midrule
    $\rvx$ & $\rvv$
      & \textbf{0.176}$^{\pm.007}$ & \textbf{0.452}$^{\pm.007}$ & \textbf{0.677}$^{\pm.006}$ & \textbf{0.803}$^{\pm.005}$ & \textbf{2.692}$^{\pm.020}$ & 11.081$^{\pm.092}$
      & 1.830$^{\pm.004}$ & 2557$^{\pm10}$ & \textbf{39169}$^{\pm169}$ & 0.004$^{\pm.000}$ \\
    $\rvx$ & $\rvepsilon$
      & 0.203$^{\pm.009}$ & 0.402$^{\pm.007}$ & 0.617$^{\pm.006}$ & 0.745$^{\pm.003}$ & 3.077$^{\pm.029}$ & 11.191$^{\pm.102}$
      & \textbf{1.673}$^{\pm.004}$ & \textbf{2530}$^{\pm9}$ & 39870$^{\pm155}$ & \textbf{0.002}$^{\pm.000}$ \\
    $\rvx$ & $\rvx$
      & 0.196$^{\pm.010}$ & 0.445$^{\pm.006}$ & 0.672$^{\pm.005}$ & 0.795$^{\pm.004}$ & 2.797$^{\pm.020}$ & \textbf{10.959}$^{\pm.086}$
      & 2.326$^{\pm.005}$ & 2407$^{\pm7}$ & 31665$^{\pm119}$ & 0.009$^{\pm.000}$ \\
    \bottomrule
  \end{tabular}
\end{table*}

\subsection{Model Sizes}
\begin{figure*}[t]
  \centering
  \begin{subfigure}{\linewidth}
    \centering
    \includegraphics[width=\linewidth]{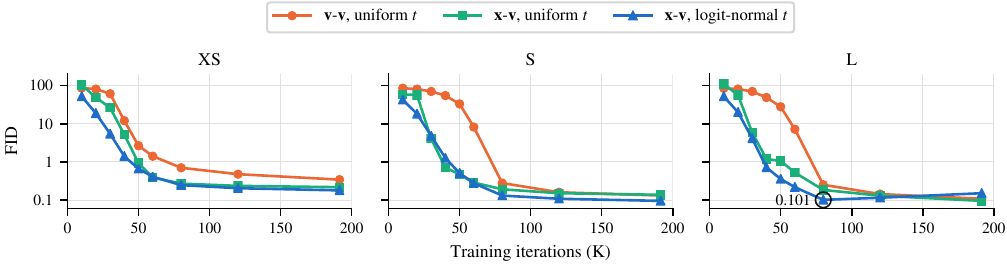}
    \caption{FID$\downarrow$}
    \label{fig:size-curves-fid}
  \end{subfigure}
  \\[4pt]
  \begin{subfigure}{\linewidth}
    \centering
    \includegraphics[width=\linewidth]{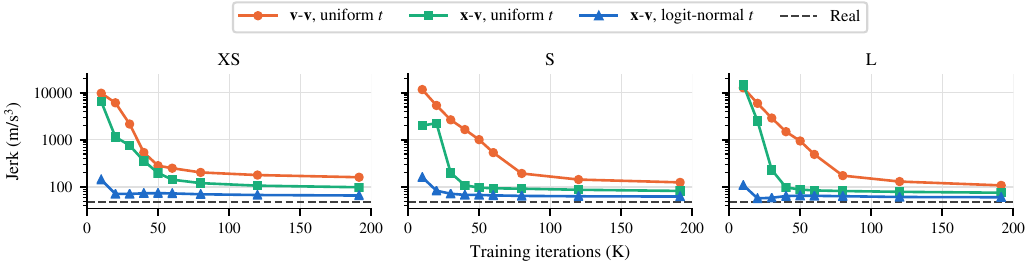}
    \caption{Jerk$\rightarrow$; the dashed line is the jerk of the real test motions.}
    \label{fig:size-curves-jerk}
  \end{subfigure}
  \caption{Training curves of the XS (4 layers, 5M parameters), S (main model, 8 layers, 39M parameters), and L (12 layers, 130M parameters) models on HumanML3D.
  Each point is a checkpoint sampled with 50 Euler steps and evaluated with 5 repeats; the learning rate is decayed by $10\times$ at 50K iterations.
  The circle marks the lowest FID of the L model with logit-normal $t$ (80K iterations), after which its FID increases.}
  \label{fig:size-curves}
\end{figure*}

\Cref{fig:size-curves} shows the FID and Jerk training curves of XS (4 layers, 4 heads, 5M parameters), S (8 layers, 8 heads, 39M parameters), and L (12 layers, 12 heads, 130M parameters) models summarized in \cref{tab:hml3d-size-fid-jerk} of the paper. We can see that the L $\rvv$-$\rvv$ model converges on FID using the full training iterations while on Jerk, and at almost 200K iterations, L $\rvv$-$\rvv$ still has higher Jerk than XS $\rvx$-$\rvv$ trained for 50K iterations. 

\subsection{Bottleneck}
\begin{figure*}[t]
  \centering
  \includegraphics[width=0.95\linewidth]{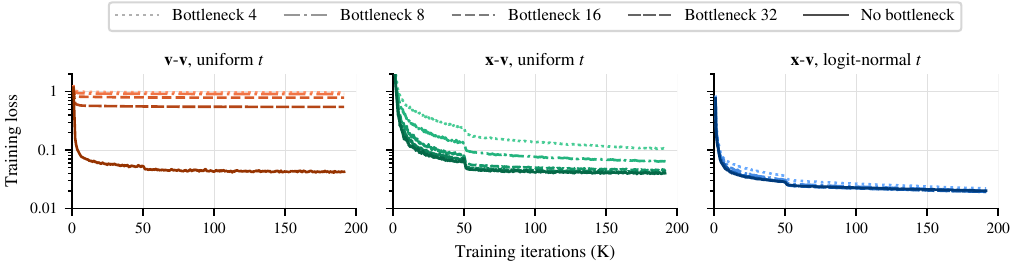}
  \caption{Training curves for $\rvv$-$\rvv$, $\rvx$-$\rvv$ and $\rvx$-$\rvv$,logit-normal t models on the four bottleneck size $d \in \{4, 8, 16, 32\}$ vs. the full-rank original runs (No bottleneck)}
  \label{fig:training-curves-bottleneck}
\end{figure*}

\Cref{tab:hml3d-bottleneck} reports the FID and Jerk of every bottleneck width plotted in \cref{fig:hml3d-bottleneck-fid-jerk} of the paper. \Cref{fig:training-curves-bottleneck} shows the training curves. The $\rvv$-$\rvv$ model's training loss with any bottleneck stays an order of magnitude above that of the full-rank model, whereas the $\rvx$-$\rvv$ models converge close to their full-rank loss, with the gap shrinking as $d$ grows and almost no gap with the logit-normal $t$-sampler. For this reason, we do not report $\rvv$-$\rvv$ results with bottlenecks in the paper.

\begin{table}[t]
  \caption{Effect of a low-rank bottleneck in the patch embedding on HumanML3D (S models, 191.5K iterations).
  The $66$-dimensional frame is embedded through a linear layer of the given width before being projected to the Transformer width; None is the standard full-rank embedding.
  We report the mean over 20 evaluation repeats.
  With a bottleneck of any width the $\rvv$-$\rvv$ model fails to train (its training loss stays at 0.55--0.99, against 0.04 without the bottleneck), so its scores are shown in grey.
  $\downarrow$: lower is better, $\rightarrow$: closer to the real test motions (48.9\,m/s$^3$) is better.}
  \label{tab:hml3d-bottleneck}
  \centering
  \footnotesize
  \setlength{\tabcolsep}{4pt}
  \begin{tabular}{@{}lccccc@{}}
    \toprule
    & \multicolumn{5}{c}{Bottleneck width} \\
    \cmidrule(l){2-6}
    Model & 4 & 8 & 16 & 32 & None \\
    \midrule
    \multicolumn{6}{@{}l}{\emph{FID$\downarrow$}} \\
    $\rvv$-$\rvv$, uniform $t$ & \textcolor{gray}{80.8} & \textcolor{gray}{80.9} & \textcolor{gray}{82.9} & \textcolor{gray}{86.9} & 0.149 \\
    $\rvx$-$\rvv$, uniform $t$ & 0.858 & 0.193 & 0.124 & 0.142 & 0.134 \\
    $\rvx$-$\rvv$, logit-normal $t$ & 1.464 & 0.235 & 0.106 & 0.102 & \textbf{0.087} \\
    \midrule
    \multicolumn{6}{@{}l}{\emph{Jerk$\rightarrow$ (m/s$^3$)}} \\
    $\rvv$-$\rvv$, uniform $t$ & \textcolor{gray}{37435} & \textcolor{gray}{36184} & \textcolor{gray}{33284} & \textcolor{gray}{26008} & 120.2 \\
    $\rvx$-$\rvv$, uniform $t$ & 145.8 & 109.8 & 82.3 & 79.2 & 83.0 \\
    $\rvx$-$\rvv$, logit-normal $t$ & 98.1 & 77.6 & 65.2 & \textbf{61.4} & 62.4 \\
    \bottomrule
  \end{tabular}
\end{table}

\subsection{Other representations}
\Cref{tab:hml3d-latent-3x3,tab:hml3d-redundant-3x3} give all metrics of the nine models trained on autoencoder latents and on the 263-dimensional redundant features of HumanML3D, of which \cref{tab:hml3d-repr-fid-jerk} of the paper shows FID and Jerk for $\rvv$-$\rvv$ and $\rvx$-$\rvv$.

\begin{table*}[t]
  \caption{Prediction target \vs loss space on autoencoder latents (HumanML3D).
  Same protocol as \cref{tab:hml3d-3x3} (size S, uniform $t$-sampler, 191.5K iterations, 50 Euler steps, classifier-free guidance scale 3), but the models are trained on the latents of the causal autoencoder of ACMDM~\cite{meng2025absolute} (4 channels, $4\times$ temporal downsampling) instead of raw joint positions; the physical-plausibility metrics are computed on the decoded joints.
  The FID and Jerk columns are the AE-latent cells of \cref{tab:hml3d-repr-fid-jerk} of the paper (for $\rvv$-$\rvv$ and $\rvx$-$\rvv$).
  We report the mean and, as a superscript, the 95\% confidence interval over 20 evaluation repeats.
  $\downarrow$: lower is better, $\uparrow$: higher is better, $\rightarrow$: closer to Real is better.
  For $\rvepsilon$-prediction we report the better in FID of exact and floored sampling; $\dagger$ marks floored sampling (\cref{tab:floor-fid}).
  Bold marks the best model in each column.}
  \label{tab:hml3d-latent-3x3}
  \centering
  \footnotesize
  \setlength{\tabcolsep}{2.2pt}
  \begin{tabular}{@{}cccccccccccc@{}}
    \toprule
    & & \multicolumn{6}{c}{Text-to-Motion metrics} & \multicolumn{4}{c}{Physical plausibility} \\
    \cmidrule(lr){3-8} \cmidrule(l){9-12}
    & & & \multicolumn{3}{c}{R-Precision$\uparrow$} & & & & & & \\
    \cmidrule(lr){4-6}
    Pred. & Loss & FID$\downarrow$ & Top 1 & Top 2 & Top 3 & MM-Dist$\downarrow$ & Diversity$\rightarrow$
      & Bone$\downarrow$ & Accel.$\rightarrow$ & Jerk$\rightarrow$ & Floor$\downarrow$ \\
    & & & & & & & & mm/frame & m/s$^2$ & m/s$^3$ & mm \\
    \midrule
    \multicolumn{2}{c}{Real}
      & -- & 0.499$^{\pm.003}$ & 0.693$^{\pm.003}$ & 0.792$^{\pm.002}$ & 3.293$^{\pm.007}$ & 10.500$^{\pm.081}$
      & 0.000$^{\pm.000}$ & 2.764$^{\pm.001}$ & 48.9$^{\pm0.0}$ & 0.000$^{\pm.000}$ \\
    \midrule
    $\rvv$ & $\rvv$
      & 0.142$^{\pm.005}$ & 0.508$^{\pm.003}$ & 0.704$^{\pm.002}$ & 0.801$^{\pm.002}$ & 3.247$^{\pm.008}$ & 10.849$^{\pm.080}$
      & 1.911$^{\pm.006}$ & 2.969$^{\pm.008}$ & 62.7$^{\pm0.2}$ & 0.074$^{\pm.003}$ \\
    $\rvv$ & $\rvepsilon$
      & 0.339$^{\pm.010}$ & 0.417$^{\pm.003}$ & 0.604$^{\pm.003}$ & 0.711$^{\pm.003}$ & 3.899$^{\pm.016}$ & 10.989$^{\pm.093}$
      & \textbf{1.634}$^{\pm.005}$ & 2.930$^{\pm.010}$ & \textbf{60.3}$^{\pm0.2}$ & 0.103$^{\pm.004}$ \\
    $\rvv$ & $\rvx$
      & 0.654$^{\pm.012}$ & 0.476$^{\pm.002}$ & 0.670$^{\pm.003}$ & 0.773$^{\pm.003}$ & 3.335$^{\pm.010}$ & 10.022$^{\pm.087}$
      & 5.257$^{\pm.009}$ & 3.961$^{\pm.009}$ & 93.9$^{\pm0.2}$ & 0.273$^{\pm.007}$ \\
    \midrule
    $\rvepsilon$\floored & $\rvv$
      & \textbf{0.082}$^{\pm.005}$ & 0.507$^{\pm.003}$ & 0.700$^{\pm.003}$ & 0.795$^{\pm.002}$ & 3.262$^{\pm.011}$ & \textbf{10.678}$^{\pm.086}$
      & 2.123$^{\pm.006}$ & 2.973$^{\pm.010}$ & 63.1$^{\pm0.2}$ & 0.070$^{\pm.006}$ \\
    $\rvepsilon$\floored & $\rvepsilon$
      & 0.204$^{\pm.007}$ & 0.426$^{\pm.002}$ & 0.616$^{\pm.002}$ & 0.722$^{\pm.002}$ & 3.810$^{\pm.012}$ & 10.688$^{\pm.090}$
      & 2.070$^{\pm.008}$ & 3.074$^{\pm.009}$ & 65.3$^{\pm0.2}$ & 0.107$^{\pm.007}$ \\
    $\rvepsilon$\floored & $\rvx$
      & 1.191$^{\pm.021}$ & 0.472$^{\pm.003}$ & 0.663$^{\pm.003}$ & 0.764$^{\pm.003}$ & 3.383$^{\pm.010}$ & 9.634$^{\pm.085}$
      & 5.713$^{\pm.010}$ & 3.948$^{\pm.008}$ & 93.4$^{\pm0.2}$ & 0.166$^{\pm.008}$ \\
    \midrule
    $\rvx$ & $\rvv$
      & 0.172$^{\pm.006}$ & \textbf{0.513}$^{\pm.003}$ & \textbf{0.706}$^{\pm.003}$ & \textbf{0.803}$^{\pm.002}$ & \textbf{3.238}$^{\pm.010}$ & 10.904$^{\pm.096}$
      & 2.084$^{\pm.006}$ & 3.088$^{\pm.010}$ & 66.2$^{\pm0.2}$ & 0.050$^{\pm.002}$ \\
    $\rvx$ & $\rvepsilon$
      & 0.314$^{\pm.009}$ & 0.443$^{\pm.002}$ & 0.632$^{\pm.002}$ & 0.736$^{\pm.002}$ & 3.707$^{\pm.009}$ & 10.959$^{\pm.098}$
      & 1.869$^{\pm.007}$ & \textbf{2.930}$^{\pm.010}$ & 61.6$^{\pm0.2}$ & 0.078$^{\pm.004}$ \\
    $\rvx$ & $\rvx$
      & 0.163$^{\pm.007}$ & 0.486$^{\pm.003}$ & 0.681$^{\pm.002}$ & 0.781$^{\pm.002}$ & 3.338$^{\pm.008}$ & 10.685$^{\pm.085}$
      & 2.776$^{\pm.007}$ & 3.346$^{\pm.009}$ & 75.3$^{\pm0.2}$ & \textbf{0.049}$^{\pm.001}$ \\
    \bottomrule
  \end{tabular}
\end{table*}

\begin{table*}[t]
  \caption{Prediction target \vs loss space on the redundant representation (HumanML3D).
  Same protocol as \cref{tab:hml3d-3x3} of the paper (size S, uniform $t$-sampler, 191.5K iterations, 50 Euler steps, classifier-free guidance scale 3), but the models are trained on the 263-dimensional redundant features of HumanML3D~\cite{guo2022generating} instead of raw joint positions; all metrics are computed on the joint positions recovered from the generated features.
  We report the mean and, as a superscript, the 95\% confidence interval over 20 evaluation repeats.
  $\downarrow$: lower is better, $\uparrow$: higher is better, $\rightarrow$: closer to Real is better.
  For $\rvepsilon$-prediction we report the better in FID of exact and floored sampling; $\dagger$ marks floored sampling (\cref{tab:floor-fid}).
  Bold marks the best model in each column.}
  \label{tab:hml3d-redundant-3x3}
  \centering
  \footnotesize
  \setlength{\tabcolsep}{2.2pt}
  \begin{tabular}{@{}cccccccccccc@{}}
    \toprule
    & & \multicolumn{6}{c}{Text-to-Motion metrics} & \multicolumn{4}{c}{Physical plausibility} \\
    \cmidrule(lr){3-8} \cmidrule(l){9-12}
    & & & \multicolumn{3}{c}{R-Precision$\uparrow$} & & & & & & \\
    \cmidrule(lr){4-6}
    Pred. & Loss & FID$\downarrow$ & Top 1 & Top 2 & Top 3 & MM-Dist$\downarrow$ & Diversity$\rightarrow$
      & Bone$\downarrow$ & Accel.$\rightarrow$ & Jerk$\rightarrow$ & Floor$\downarrow$ \\
    & & & & & & & & mm/frame & m/s$^2$ & m/s$^3$ & mm \\
    \midrule
    \multicolumn{2}{c}{Real}
      & -- & 0.497$^{\pm.002}$ & 0.694$^{\pm.002}$ & 0.794$^{\pm.002}$ & 3.290$^{\pm.008}$ & 10.582$^{\pm.102}$
      & 0.000$^{\pm.000}$ & 2.764$^{\pm.001}$ & 48.9$^{\pm0.0}$ & 0.000$^{\pm.000}$ \\
    \midrule
    $\rvv$ & $\rvv$
      & 0.399$^{\pm.011}$ & 0.482$^{\pm.003}$ & 0.679$^{\pm.002}$ & 0.778$^{\pm.002}$ & 3.411$^{\pm.009}$ & 11.035$^{\pm.071}$
      & 0.613$^{\pm.003}$ & \textbf{2.732}$^{\pm.009}$ & 52.3$^{\pm0.2}$ & 0.114$^{\pm.005}$ \\
    $\rvv$ & $\rvepsilon$
      & 1.876$^{\pm.028}$ & 0.206$^{\pm.002}$ & 0.343$^{\pm.003}$ & 0.448$^{\pm.003}$ & 5.497$^{\pm.011}$ & 9.436$^{\pm.073}$
      & \textbf{0.522}$^{\pm.003}$ & 3.304$^{\pm.014}$ & 64.3$^{\pm0.3}$ & 0.172$^{\pm.003}$ \\
    $\rvv$ & $\rvx$
      & \textbf{0.199}$^{\pm.007}$ & \textbf{0.497}$^{\pm.003}$ & 0.689$^{\pm.002}$ & 0.788$^{\pm.003}$ & \textbf{3.288}$^{\pm.009}$ & 10.826$^{\pm.091}$
      & 6.772$^{\pm.009}$ & 7.393$^{\pm.011}$ & 244.5$^{\pm0.3}$ & 0.127$^{\pm.005}$ \\
    \midrule
    $\rvepsilon$\floored & $\rvv$
      & 1.230$^{\pm.028}$ & 0.400$^{\pm.003}$ & 0.585$^{\pm.003}$ & 0.691$^{\pm.003}$ & 3.937$^{\pm.016}$ & 10.000$^{\pm.089}$
      & 1.453$^{\pm.002}$ & 3.055$^{\pm.006}$ & 74.6$^{\pm0.2}$ & \textbf{0.056}$^{\pm.001}$ \\
    $\rvepsilon$\floored & $\rvepsilon$
      & 6.395$^{\pm.052}$ & 0.289$^{\pm.003}$ & 0.453$^{\pm.003}$ & 0.564$^{\pm.002}$ & 4.756$^{\pm.013}$ & 8.159$^{\pm.072}$
      & 10.43$^{\pm.02}$ & 11.65$^{\pm.02}$ & 397.9$^{\pm0.7}$ & 0.385$^{\pm.003}$ \\
    $\rvepsilon$ & $\rvx$
      & 0.753$^{\pm.017}$ & 0.437$^{\pm.003}$ & 0.623$^{\pm.002}$ & 0.726$^{\pm.002}$ & 3.719$^{\pm.014}$ & \textbf{10.347}$^{\pm.078}$
      & 9.295$^{\pm.231}$ & 10.48$^{\pm.32}$ & 351.9$^{\pm11.6}$ & 11.16$^{\pm.37}$ \\
    \midrule
    $\rvx$ & $\rvv$
      & 0.513$^{\pm.012}$ & 0.451$^{\pm.002}$ & 0.647$^{\pm.002}$ & 0.750$^{\pm.002}$ & 3.600$^{\pm.011}$ & 11.080$^{\pm.069}$
      & 0.865$^{\pm.003}$ & 3.329$^{\pm.009}$ & 81.1$^{\pm0.2}$ & 0.088$^{\pm.005}$ \\
    $\rvx$ & $\rvepsilon$
      & 2.288$^{\pm.030}$ & 0.125$^{\pm.002}$ & 0.224$^{\pm.002}$ & 0.306$^{\pm.002}$ & 6.534$^{\pm.013}$ & 9.118$^{\pm.076}$
      & 0.972$^{\pm.007}$ & 3.597$^{\pm.014}$ & 81.5$^{\pm0.3}$ & 0.151$^{\pm.005}$ \\
    $\rvx$ & $\rvx$
      & 0.272$^{\pm.009}$ & \textbf{0.497}$^{\pm.004}$ & \textbf{0.692}$^{\pm.002}$ & \textbf{0.791}$^{\pm.002}$ & 3.314$^{\pm.009}$ & 10.978$^{\pm.087}$
      & 0.824$^{\pm.003}$ & 2.813$^{\pm.009}$ & \textbf{52.2}$^{\pm0.2}$ & 0.063$^{\pm.003}$ \\
    \bottomrule
  \end{tabular}
\end{table*}

\subsection{Flooring at Sampling}
\Cref{tab:hml3d-sampling-floor-full} is the full version of \cref{tab:hml3d-sampling-floor} at \cref{supp:flooring}.
Flooring the conversion when sampling $\rvx$-$\rvv$ degrades every T2M and plausibility metric, with both the uniform and the logit-normal $t$-sampler, and only changes floor penetration marginally.

\begin{table*}[t]
  \caption{Full results of \cref{tab:hml3d-sampling-floor}: effect of flooring the conversion to velocity when sampling $\rvx$-prediction models on HumanML3D, with the exact conversion and with its denominator $1-t$ floored at $\delta=0.05$.
  We report the mean and, as a superscript, the 95\% confidence interval over 20 evaluation repeats.
  $\downarrow$: lower is better, $\uparrow$: higher is better, $\rightarrow$: closer to Real is better.
  Bold marks the better of the two samplers for each model.}
  \label{tab:hml3d-sampling-floor-full}
  \centering
  \footnotesize
  \setlength{\tabcolsep}{2.8pt}
  \begin{tabular}{@{}lcccccccccc@{}}
    \toprule
    & \multicolumn{6}{c}{Text-to-Motion metrics} & \multicolumn{4}{c}{Physical plausibility} \\
    \cmidrule(lr){2-7} \cmidrule(l){8-11}
    & & \multicolumn{3}{c}{R-Precision$\uparrow$} & & & & & & \\
    \cmidrule(lr){3-5}
    Sampler & FID$\downarrow$ & Top 1 & Top 2 & Top 3 & MM-Dist$\downarrow$ & Diversity$\rightarrow$
      & Bone$\downarrow$ & Accel.$\rightarrow$ & Jerk$\rightarrow$ & Floor$\downarrow$ \\
    & & & & & & & mm/frame & m/s$^2$ & m/s$^3$ & mm \\
    \midrule
    Real
      & -- & 0.500$^{\pm.002}$ & 0.694$^{\pm.003}$ & 0.793$^{\pm.002}$ & 3.293$^{\pm.008}$ & 10.506$^{\pm.105}$
      & 0.000$^{\pm.000}$ & 2.764$^{\pm.001}$ & 48.9$^{\pm0.0}$ & 0.000$^{\pm.000}$ \\
    \midrule
    \multicolumn{11}{@{}l}{\emph{$\rvx$-$\rvv$, uniform $t$}} \\
    Exact
      & \textbf{0.134}$^{\pm.004}$ & \textbf{0.508}$^{\pm.003}$ & \textbf{0.700}$^{\pm.003}$ & \textbf{0.797}$^{\pm.003}$ & \textbf{3.251}$^{\pm.010}$ & \textbf{10.803}$^{\pm.075}$
      & \textbf{1.573}$^{\pm.003}$ & \textbf{3.646}$^{\pm.010}$ & \textbf{83.0}$^{\pm0.2}$ & \textbf{0.055}$^{\pm.003}$ \\
    Floored
      & 12.198$^{\pm.050}$ & 0.362$^{\pm.002}$ & 0.535$^{\pm.003}$ & 0.640$^{\pm.002}$ & 4.598$^{\pm.010}$ & 7.805$^{\pm.058}$
      & 15.008$^{\pm.003}$ & 16.345$^{\pm.006}$ & 576.3$^{\pm0.1}$ & 0.063$^{\pm.002}$ \\
    \midrule
    \multicolumn{11}{@{}l}{\emph{$\rvx$-$\rvv$, logit-normal $t$}} \\
    Exact
      & \textbf{0.087}$^{\pm.006}$ & \textbf{0.475}$^{\pm.003}$ & \textbf{0.665}$^{\pm.003}$ & \textbf{0.766}$^{\pm.002}$ & \textbf{3.447}$^{\pm.012}$ & \textbf{10.507}$^{\pm.073}$
      & \textbf{1.964}$^{\pm.003}$ & \textbf{3.070}$^{\pm.006}$ & \textbf{62.4}$^{\pm0.1}$ & 0.065$^{\pm.001}$ \\
    Floored
      & 15.069$^{\pm.065}$ & 0.333$^{\pm.002}$ & 0.492$^{\pm.002}$ & 0.596$^{\pm.002}$ & 4.966$^{\pm.011}$ & 7.383$^{\pm.056}$
      & 14.998$^{\pm.002}$ & 15.923$^{\pm.003}$ & 563.4$^{\pm0.1}$ & \textbf{0.057}$^{\pm.001}$ \\
    \bottomrule
  \end{tabular}
\end{table*}

\end{document}